\documentclass{article} 
\usepackage{collas2022_conference,times}

\usepackage{hyperref}
\hypersetup{
    colorlinks=true,
    linkcolor=red,
    filecolor=magenta,      
    urlcolor=blue,
    citecolor=purple,
    pdftitle={Overleaf Example},
    pdfpagemode=FullScreen,
    }

\usepackage{wrapfig}
\usepackage{listings}

\usepackage[]{multicol}
\usepackage{multirow}
\usepackage{cite}
\usepackage{comment}
\usepackage[algo2e]{algorithm2e}
\usepackage{soul}
\usepackage{caption, subcaption}
\usepackage[noend]{algorithmic}
\usepackage{algorithm}

\newtheorem{mydef}{Definition}

\newtheorem{proof}{Proof}

\newtheorem{theorem}{Theorem}

\usepackage{amsmath,amssymb,amsfonts}
\usepackage{adjustbox}
\usepackage{balance}
\usepackage{hyperref}
\usepackage{bm}
\usepackage{graphicx}
\usepackage{textcomp}
\usepackage{cite}
\usepackage{diagbox}
\usepackage{makecell}

\title{Lifelong DP: Consistently Bounded \\ Differential Privacy in Lifelong Machine Learning}

\author{Phung Lai, Han Hu, NhatHai Phan\thanks{Corresponding Author.} \\
New Jersey Insitute of Technology, USA\\
\texttt{\{tl353,hh255,phan\}@njit.edu} 
\And 
Ruoming Jin \\
Kent State University, USA \\
\texttt{\{rjin1\}@kent.edu} 
\AND 
My T. Thai \\
University of Florida, USA \\
\texttt{\{mythai\}@cise.ufl.edu} \\
\And
An M. Chen \\
Qualcomm Incorporated, USA \\
\texttt{\{anc\}@qualcomm.com}
}

\collasfinalcopy 

\begin{document}

\maketitle

\begin{abstract}
In this paper, we show that the process of continually learning new tasks and memorizing previous tasks introduces unknown privacy risks and challenges to bound the privacy loss. Based upon this, we introduce a formal definition of \textbf{Lifelong DP}, in which the participation of any data tuples in the training set of any tasks is protected, under a consistently bounded DP protection, given a growing stream of tasks. A consistently bounded DP means having only one fixed value of the DP privacy budget, regardless of the number of tasks. To preserve Lifelong DP, we propose a scalable and heterogeneous algorithm, called \textbf{L2DP-ML} with a streaming batch training, to efficiently train and continue releasing new versions of an L2M model, given the heterogeneity in terms of data sizes and the training order of tasks, without affecting DP protection of the private training set. An end-to-end theoretical analysis and thorough evaluations show that our mechanism is significantly better than baseline approaches in preserving Lifelong DP. The implementation of L2DP-ML is available at: \url{https://github.com/haiphanNJIT/PrivateDeepLearning}.
\end{abstract}

\vspace{-5pt}
\section{Introduction}
\vspace{-5pt}

Lifelong learning (L2M) is crucial for machine learning (ML) to acquire new skills through continual learning, pushing ML toward a more human learning in reality. Given a stream of different tasks and data, a deep neural network (DNN) can quickly learn a new task, by leveraging the acquired knowledge after learning previous tasks, under constraints in terms of the amount of computing and memory required \citep{chaudhry2018efficient}. As a result, it is quite challenging to train an L2M model with a high utility.
Orthogonal to this, L2M models are vulnerable to adversarial attacks, i.e., privacy model attacks \citep{2016arXiv161005820S,Fredrikson:2015:MIA,DBLP:conf/ijcai/SW15,DBLP:journals/corr/PapernotMSW16}, when DNNs are trained on highly sensitive data, e.g., clinical records \citep{Choiocw112,citeulike:14040136}, user profiles \citep{Roumia2014,citeulike:7685411}, and medical images \citep{10.3389/fnins.2014.00229,helmstaedter2013}. 

In practice, the privacy risk will be more significant since an adversary can observe multiple versions of an L2M model released after training on each task. Different versions of the model parameters can be considered as an additional information leakage, compared with a model trained on a single task (Theorem \ref{PRTheorem}). Memorizing previous tasks while learning new tasks further exposes private information in the training set, by continuously accessing the data from the previously learned tasks (i.e., data stored in an episodic memory \citep{chaudhry2018efficient,riemer2018learning,tao2020fewshot}); or accessing adversarial examples produced from generative memories to imitate real examples of past tasks \citep{10.5555/3294996.3295059,NIPS2018_7836,DBLP:conf/cvpr/OstapenkoPKJN19}.
Unfortunately, there is a lack of study offering privacy protection to the training data in L2M. 

\textbf{Our Contributions.} To address this problem, we propose to preserve differential privacy (DP) \citep{dwork2006calibrating}, a rigorous formulation of privacy in probabilistic terms, in L2M. We introduce a new definition of lifelong differential privacy (Lifelong DP), in which the participation of any data tuple in any tasks is protected under a \textit{consistently bounded} DP guarantee, given the released parameters in both learning new tasks and memorizing previous tasks (Definition \ref{LifelongDP}). 
This is significant by allowing us to train and release new versions of an L2M model, given a stream of tasks and data, under DP protection.

Based upon this, we propose a novel \textsc{L2DP-ML} algorithm to preserve Lifelong DP. In \textsc{L2DP-ML}, privacy-preserving noise is injected into inputs and hidden layers to achieve DP in learning private model parameters in each task (Alg. \ref{DPL2M-Dataset}). Then, we configure the episodic memory as a stream of fixed and disjoint batches of data, to efficiently achieve Lifelong DP (Theorem \ref{OverallL2DP}). The previous task memorizing constraint is solved, by inheriting the recipe of the well-known A-gem algorithm \citep{chaudhry2018efficient}, under Lifelong DP.
To our knowledge, our study establishes a formal connection between DP preservation and L2M given a growing number of learning tasks compared with existing works \citep{farquhar2019differentially, phan2019private}. 
Rigorous experiments, conducted on permuted MNIST \citep{Kirkpatrick3521}, permuted CIFAR-10 datasets, and an L2M task on our collected dataset for human activity recognition in the wild show promising results in preserving DP in L2M. 


\vspace{-5pt}
\section{Background}
\vspace{-5pt}

Let us first revisit L2M with A-gem and DP.
In L2M, we learn a sequence of tasks $\mathbb{T} = \{t_1, \ldots, t_m\}$ one by one, such that the learning of each new task will not forget the models learned for the previous tasks.
Let $D_i$ be the dataset of the $i$-th task. Each tuple contains data $x \in [-1, 1]^d$ and a ground-truth label $y \in \mathbb{Z}_K$, which is a one-hot vector of $K$ categorical outcomes $y = \{y_{1}, \ldots, y_{K}\}$. A single true class label $y_x \in y$ given $x$ is assigned to only one of the $K$ categories. All the training sets $D_i$ are non-overlapping; that is, an arbitrary input $(x, y)$ belongs to only one $D_i$, i.e., $\exists! i \in [1, m]: (x, y) \in D_i$ ($x \in D_i$ for simplicity).
On input $x$ and parameters $\theta$, a model outputs class scores $f: \mathbb{R}^d \rightarrow \mathbb{R}^K$ that map inputs $x$ to a vector of scores $f(x) = \{f_1(x), \ldots, f_K(x)\}$ s.t. $\forall k \in [1, K]: f_k(x) \in [0, 1]$ and $\sum_{k = 1}^K f_k(x) = 1$. The class with the highest score is selected as the predicted label for $x$, denoted as $y(x) = \max_{k \in K} f_k(x)$. 
A loss function $L(f(\theta, x), y)$ presents the penalty for mismatching between the predicted values $f(\theta, x)$ and original values $y$.

\textbf{Lifelong Learning.} Given the current task $\tau$ $(\leq m)$, let us denote $\mathbb{T}_\tau = \{t_1, \ldots, t_{\tau-1}\}$ is a set of tasks that have been learnt. Although there are different L2M settings, i.e., episodic memory \citep{8100070,NIPS2017_7225,riemer2018learning,abati2020conditional,tao2020fewshot,rajasegaran2020itaml,ebrahimi2020adversarial} and generative memory \citep{10.5555/3294996.3295059,NIPS2018_7836,DBLP:conf/cvpr/OstapenkoPKJN19}, we leverage one of the state-of-the-art algorithms, i.e., A-gem \citep{chaudhry2018efficient}, to demonstrate our privacy-preserving mechanism, without loss of the generality of our study. 
A-gem avoids catastrophic forgetting by storing an episodic memory $M_i$ for each task $t_i \in \mathbb{T}_\tau$. When minimizing the loss on the current task $\tau$, a typical approach is to treat the losses on the episodic memories of tasks $i < \tau$, given by $L(f(\theta, M_i)) = \frac{1}{\lvert M_i \rvert} \sum_{x \in M_i} L(f(\theta, x), y)$, as inequality constraints. In A-gem, the L2M objective function is: 
\begin{align}
 \theta^{\tau} = \arg\min_{\theta} L\big(f(\theta, D_\tau)\big) \text{ \ \ \ s.t. \ \ \ } L\big(f(\theta^\tau, \mathbb{M}_\tau)\big) \leq L\big(f(\theta^{\tau - 1}, \mathbb{M}_\tau)\big)  \label{AgemObjF}
\end{align}
where $\theta^{\tau - 1}$ are the values of model parameters $\theta$ learned after training the task $t_{\tau-1}$, $\mathbb{M}_\tau = \cup_{i < \tau} M_i$ is the episodic memory with $\mathbb{M}_1 = \emptyset$, $L\big(f(\theta^{\tau - 1}, \mathbb{M}_\tau)\big) = \sum_{i = 1}^{\tau-1} L\big(f(\theta^{\tau - 1}, M_i)\big) / (\tau - 1)$. Eq. \ref{AgemObjF} indicates that learning 
$\theta^\tau$ given the task $\tau$ will not forget previously learned tasks $\{t_1, \ldots, t_{\tau - 1}\}$ enforced by the memory replaying constraint $L\big(f(\theta^\tau, \mathbb{M}_\tau)\big) \leq L\big(f(\theta^{\tau - 1}, \mathbb{M}_\tau)\big)$.

At each training step, A-gem \citep{chaudhry2018efficient} has access to only $D_\tau$ and $\mathbb{M}_\tau$ to compute the \textit{projected gradient} $\tilde{g}$ (i.e., by addressing the constraint in Eq. \ref{AgemObjF}), as follows:
\begin{equation}
\tilde{g} = g - \frac{g^{\top}g_{ref}}{g^{\top}_{ref}g_{ref}}g_{ref}
\label{agem} 
\end{equation} 
where $g$ is the \textit{updated gradient} computed on a batch sampled from $D_\tau$, $g_{ref}$ is an \textit{episodic gradient} computed on a batch sampled from $\mathbb{M}_\tau$, and $\tilde{g}$ is used to update the model parameters $\theta$ in Eq. \ref{AgemObjF}.

\textbf{Differential Privacy (DP).} DP guarantees that the released statistical results, computed from the underlying sensitive data, is insensitive to the presence or absence of one tuple  in a dataset. Let us briefly revisit the definition of DP, as: 
\begin{mydef}{$(\epsilon, \delta)$-DP \citep{dwork2006calibrating}.} A randomized algorithm $A$ is $(\epsilon, \delta)$-DP, if for any two neighboring databases $D$ and $D'$ differing at most one tuple, and $\forall O \subseteq Range(A)$, we have:
\begin{equation}
Pr[A(D) = O] \leq e^\epsilon Pr[A(D') = O] + \delta 
\end{equation}
where $\epsilon$ controls the amount by which the distributions induced by $D$ and $D'$ may differ, and $\delta$ is a broken probability. A smaller $\epsilon$ enforces a stronger privacy guarantee. 
\label{Different Privacy} 
\end{mydef}


DP has been preserved in many ML models and tasks \citep{abadi2017protection,PhanIJCAI,papernot2018scalable}. However,
 existing mechanisms have not been designed to preserve DP in L2M under a fixed and consistently bounded privacy budget given a growing stream of learning tasks. That differs from our goal in this study.

\section{Privacy Risk and Lifelong DP}
\label{Privacy Risk - Problem Statement}
\vspace{-5pt}

In this section, we focus on analyzing the unknown privacy risk in L2M and introduce a new concept of Lifelong DP.

\textbf{Privacy Risk Analysis.} 
One benefit of L2M is that end-users can use an L2M model after training each task $\tau$, instead of waiting for the model to be trained on all the tasks. Thus, in practice, the adversary can observe the model parameters $\theta^1, \ldots, \theta^{m}$ after training each task $t_1, \ldots, t_m$. Note that the adversary does not observe any information about the (black-box) training algorithm.
Another key property in an L2M model is the episodic memory, which is kept to be read at each training step incurring privacy leakage. Therefore, the training data $D$ and episodic memory $M$ need to be protected together across tasks. Finally, in L2M, at each training step for any task $t_i$ $(i \in [1, m])$, we only have access to $D_i$ and $\mathbb{M}_i$, without a complete view of the cumulative dataset of all the tasks $\cup_{i \in [1, m]} D_i$ and $\mathbb{M}_m = \cup_{i \in [1, m -1]} M_i$. This is different from the traditional definition of a database in both DP (Def. \ref{Different Privacy}) and in a model trained on a single task. To cope with this, we propose a new definition of lifelong neighboring databases, as follows: 
\begin{mydef}{Lifelong Neighboring Databases.} Given any two lifelong databases $\mathsf{data}_m = \{\mathcal{D}, \mathcal{M}\}$ and $\mathsf{data}'_m = \{\mathcal{D}', \mathcal{M}'\}$, where $\mathcal{D} = \{D_1, \ldots, D_m\}$, $\mathcal{D}' = \{D'_1, \ldots, D'_m\}$, $\mathcal{M} = \{\mathbb{M}_1, \ldots, \mathbb{M}_m\}$, $\mathcal{M}' = \{\mathbb{M}'_1, \ldots, \mathbb{M}'_m\}$, $\mathbb{M}_i = \cup_{j \in [1, i - 1]} M_j$, and $\mathbb{M}'_i = \cup_{j \in [1, i - 1]} M'_j$. $\mathsf{data}_m$ and $\mathsf{data}'_m$ are called lifelong neighboring databases if, $\forall i \in [1, m]$: \textbf{(1)} $D_i$ and $D'_i$ differ at most one tuple; and \textbf{(2)} $M_i$ and $M'_i$ differ at most one tuple.
\label{L2ND}
\end{mydef}

\textbf{A Naive Mechanism.} To preserve DP in L2M, one can employ the moments accountant  \citep{Abadi} to train the model $f$ by injecting Gaussian noise into clipped gradients $g$ and $g_{ref}$ (Eq. \ref{agem}), with privacy budgets $\epsilon_{D_\tau}$ and $\epsilon_{\mathbb{M}_\tau}$  on each dataset $D_\tau$ and on the episodic memory $\mathbb{M}_\tau$, and a gradient clipping bound $C$. 
The post-processing property in DP \citep{Dwork:2014:AFD:2693052.2693053} can be applied to guarantee that $\tilde{g}$, computed from the perturbed $g$ and $g_{ref}$, is also DP. 

Let us denote this mechanism as $A$, and denote $A_\tau$ as $A$ applied on the task $\tau$. 
A naive approach \citep{desai2021continual} is to repeatedly apply $A$ on the sequence of tasks $\mathbb{T}$. 
Since training data is non-overlapping among tasks, the parallel composition property in DP \citep{10.1145/1536414.1536466} can be applied to estimate the total privacy budget consumed across all the tasks, as follows: 
\begin{equation}
Pr[A(\mathsf{data}_m) = \{\theta^i\}_{i \in [1, m]}] \leq  e^\epsilon Pr[A(\mathsf{data}'_m) = \{\theta^i\}_{i \in [1, m]}]  + \delta
\label{PR2}
\end{equation}
where $\epsilon = \max_{i \in [1, m]} (\epsilon_{D_i} + \epsilon_{\mathbb{M}_i})$,  and $\forall i, j \in [1, m]: \delta$ is the same for $\epsilon_{D_i}$ and $\epsilon_{\mathbb{M}_j}$.

$A(\mathsf{data}_m)$ indicates that the model is trained from scratch with the mechanism $A$, given randomly initiated parameters $\theta^0$, i.e., $A(\theta^0, \mathsf{data}_m)$.
Intuitively, we can achieve the traditional DP guarantee in L2M, as the participation of a particular data tuple in each dataset $D_\tau$ is protected under the released $(\epsilon, \delta)$-DP $\{\theta^i\}_{i \in [1, m]}$.
However, this approach introduces unknown privacy risks in each task and in the whole training process, as discussed next.


Observing the intermediate parameters $\{\theta^i\}_{i<\tau}$ turns the mechanism $A_\tau$ into a list of adaptive DP mechanisms $A_1, \ldots, A_{\tau}$ sequentially applied on tasks $t_1, \ldots, t_\tau$, where $A_i: (\prod_{j = 1}^{i - 1} \mathcal{R}_j) \times D_i \rightarrow \mathcal{R}_i$. This is an instance of adaptive composition, which we can model by using the output of all the previous mechanisms $\{\theta^i\}_{i<\tau}$ as the auxiliary input of the $A_\tau$ mechanism. Thus, given an outcome $\theta^\tau$, the privacy loss $c(\cdot)$ at $\theta^\tau$ can be measured as follows: 
\begin{equation}
c(\theta^\tau; A_\tau, \{\theta^i\}_{i<\tau}, \mathsf{data}_\tau, \mathsf{data}'_\tau)
= \log \frac{Pr[A_\tau(\{\theta^i\}_{i<\tau}, \mathsf{data}_\tau) = \theta^\tau]}{Pr[A_\tau(\{\theta^i\}_{i<\tau}, \mathsf{data}'_\tau) = \theta^\tau]}
\end{equation}



The privacy loss is accumulated across tasks, as follows:

\begin{theorem}
$
\forall \tau > 1: c(\theta^\tau; A_\tau, \{\theta^i\}_{i<\tau}, \mathsf{data}_\tau, \mathsf{data}'_\tau) =
\sum_{i = 1}^{\tau} c(\theta^i; A_i, \{\theta^j\}_{j<i}, \mathsf{data}_i, \mathsf{data}'_i).
$ 
\label{PRTheorem}
\end{theorem}



As a result of the Theorem \ref{PRTheorem}, the privacy budget at each task $\tau$ cannot be simply bounded by $\max_{\tau \in [1, m]} (\epsilon_{D_\tau} + \epsilon_{\mathbb{M}_\tau})$, given $\delta$ (Eq. \ref{PR2}).
This problem might be addressed by replacing the $\max$ function in Eq. \ref{PR2} with a summation function: $\epsilon = \sum_{\tau \in [1, m]} (\epsilon_{D_\tau} + \epsilon_{\mathbb{M}_\tau})$, 
to compute the upper bound of the privacy budget for an entire of the continual learning process. To optimize this naive approach, one can adapt the management policy  \citep{10.1145/3352020.3352032} to redistribute the privacy budget across tasks while limiting the total privacy budget $\epsilon$ to be smaller than a predefined upper bound, that is, the training will be terminated when $\epsilon$ reaches the predefined upper bound.

However, the challenge in bounding the privacy risk is still the same, centering around the growing number of tasks $m$ and the heterogeneity among tasks: \textbf{(1)} The larger the number of tasks, the larger the privacy budget will be consumed by the $\sum$ function. It is hard to identify an upper bound privacy budget given an unlimited number of streaming tasks in L2M;
\textbf{(2)} Different tasks may require different numbers of training steps due to the difference in terms of the number of tuples in each task; thus, affecting the privacy budget $\epsilon$; and \textbf{(3)} The order of training tasks also affect the privacy budget, since computing $g_{ref}$ by using data in the episodic memory from one task may be more than other tasks.
Therefore, bounding the DP budget in L2M is non-trivial. 

\textbf{Lifelong DP.}
To address these challenges, we propose a new definition of $\epsilon$-Lifelong DP to guarantee that an adversary cannot infer whether a data tuple is in the lifelong training dataset $\mathsf{data}_m$, given the released parameters $\{\theta^i\}_{i \in [1,m]}$ learned from a growing stream of an infinite number of new tasks, denoted $\forall m \in [1, \infty)$, under a consistently bounded DP budget $\epsilon $ (Eq. \ref{Cond3}). A consistently bounded DP means having only one fixed value of  $\epsilon$, regardless of the number of tasks $m$. 
In other words, it does not exist an $i \leq m$ and an $\epsilon' < \epsilon$, such that releasing $\{\theta^j\}_{j \in [1,i]}$ given training dataset $\mathsf{data}_i$ is $\epsilon' $-DP (Eq. \ref{Cond4}). 
A consistently bounded DP is significant by enabling us to keep training and releasing an L2M model without intensifying the end-to-end privacy budget consumption.
Lifelong DP can be formulated as follows: 
\begin{mydef}{$\epsilon$-Lifelong DP.} Given a lifelong database $\mathsf{data}_m$, a randomized algorithm $A$ achieves $\epsilon$-Lifelong DP, if for any of two lifelong neighboring databases $(\mathsf{data}_m, \mathsf{data}'_m)$, for all possible outputs $\{\theta^i\}_{i \in [1,m]} \in Range(A)$, $\forall m \in [1, \infty)$ we have that
\begin{align}
& P\big[A(\mathsf{data}_m) = \{\theta^i\}_{i \in [1,m]} \big] \leq e^\epsilon P\big[A(\mathsf{data}'_m) = \{\theta^i\}_{i \in [1,m]} \big] \label{Cond3} 
\\
&  \nexists (\epsilon' < \epsilon, i \leq m): P\big[A(\mathsf{data}_i) = \{\theta^j\}_{j \in [1,i]} \big] \leq  e^{\epsilon'} P\big[A(\mathsf{data}'_i) = \{\theta^j\}_{j \in [1,i]} \big] \label{Cond4} 
\end{align}
where $Range(A)$ denotes every possible output of $A$.
\label{LifelongDP}
\end{mydef} 

In our Lifelong DP definition, the episodic memory (data) $\mathcal{M}$ can be an empty set $\emptyset$ in the definition of lifelong neighboring databases (Def. \ref{L2ND}) given L2M mechanisms that do not need to access $\mathcal{M}$ \citep{yoon2020xtarnet,ye2020learning,maschler2021regularization,qu2021recent,he2022online}.

To preserve Lifelong DP, we need to address the following problems:  \textbf{(1)} The privacy loss accumulation across tasks; \textbf{(2)} The overlapping between the episodic memory $\mathcal{M}$ and the training data $\mathcal{D}$; and \textbf{(3)} The data sampling process for computing the episodic gradient $g_{ref}$ given the growing episodic memory $\mathcal{M}$. The root cause issue of these problems is that in an L2M model, the episodic memory $\mathcal{M}$, which accumulatively stores data from all of the previous tasks, is read at each training step. Thus, using the moments account to preserve Lifelong DP will cause the privacy budget accumulated, resulting in a loose privacy protection given a large number of tasks or training steps.  
Therefore, designing a mechanism to preserve Lifelong DP under a tight privacy budget is non-trivial and an open problem.

\begin{tabular}{cc}
\begin{minipage}{0.6\textwidth} 
\begin{algorithm}[H]
\footnotesize
\captionof{algorithm}{\textsc{L2DP-ML} Algorithm}
\label{DPL2M-Dataset}
\KwIn{$\epsilon_1,\epsilon_2,\mathbb{T}$=$\{t_i\}_{i \in [1,m]}, \{D_i\}_{i \in [1,m]}$}
\KwOut{ $(\epsilon_1 + \epsilon_1/\gamma_{\mathbf{x}} + \epsilon_1/\gamma+\epsilon_2)$-Lifelong DP parameters $\{\theta^i\}_{i \in [1, m]} = \{\theta^{i}_1, \theta^{i}_2\}_{i \in [1, m]}$}
\begin{algorithmic}[1]
\STATE \textbf{Draw Noise} $\chi_1 \leftarrow [Lap(\frac{\Delta_{\widetilde{\mathcal{R}}}}{\epsilon_1})]^{d}$, $\chi_2 \leftarrow [Lap(\frac{\Delta_{\widetilde{\mathcal{R}}}}{\epsilon_1})]^{\beta}$, $\chi_3 \leftarrow [Lap(\frac{\Delta_{\widetilde{\mathcal{L}}}}{\epsilon_2})]^{\lvert \mathbf{h}_\pi \lvert}$
\STATE \textbf{Randomly Initialize:} $\theta^0 = \{\theta^0_{1}, \theta^0_{2}\}$, $\mathbb{M}_1 = \emptyset$, $\forall \tau \in \mathbb{T}: \overline{D}_\tau = \{\overline{x}_r \leftarrow x_{r} + \frac{\chi_1}{\lvert D_\tau\lvert }\}_{x_r \in D_\tau}$, hidden layers $\{\mathbf{h}_1 + \frac{2\chi_2}{\lvert D_\tau\lvert },\dots, \mathbf{h}_\pi\}$
\FOR{$\tau \in [1, m]$}
        \IF{$\tau == 1$}
        	  \STATE $g \leftarrow \{\nabla_{\theta_1}\overline{\mathcal{R}}_{\overline{D}_\tau}(\theta^{\tau-1}_1), \nabla_{\theta_2} \overline{\mathcal{L}}_{\overline{D}_\tau}(\theta^{\tau-1}_2)\}$ with the noise $\frac{\chi_3}{\lvert D_\tau\lvert }$
	    \ELSE
	    	  \STATE $\mathbb{M}_{\tau} \leftarrow \mathbb{M}_{\tau-1} \cup \{\overline{D}_{\tau-1}\}$
	    	  \STATE \textbf{Randomly Pick} a dataset $\overline{D}_{ref} \in \mathbb{M}_{\tau}$
	        \STATE $g \leftarrow \{\nabla_{\theta_1}\overline{\mathcal{R}}_{\overline{D}_\tau}(\theta^{\tau-1}_1), \nabla_{\theta_2} \overline{\mathcal{L}}_{\overline{D}_\tau}(\theta^{\tau-1}_2)\}$ with the noise $\frac{\chi_3}{\lvert D_\tau\lvert }$
	        \STATE $g_{ref} \leftarrow \{\nabla_{\theta_1}\overline{\mathcal{R}}_{\overline{D}_{ref}}(\theta^{\tau-1}_1), \nabla_{\theta_2} \overline{\mathcal{L}}_{\overline{D}_{ref}}(\theta^{\tau-1}_2)\}$ with the noise $\frac{\chi_3}{\lvert D_{ref}\lvert }$
	        \STATE $\tilde{g} \leftarrow g - \frac{g^{\top}g_{ref}}{g^{\top}_{ref}g_{ref}}g_{ref}$
	    \ENDIF
	    \STATE \textbf{Descent: } $\{\theta^{\tau}_1, \theta^{\tau}_2\} \leftarrow \{\theta^{\tau-1}_1, \theta^{\tau-1}_2\} - \varrho \tilde{g}$ \#~learning rate~$\varrho$\\
	    \STATE \textbf{Release: } $\{\theta^{\tau}_1, \theta^{\tau}_2\}$
\ENDFOR 
\end{algorithmic} 
\end{algorithm} 
\end{minipage}
&
\begin{minipage}[t]{0.38\textwidth} \vspace{-120pt}
  \centering
  \includegraphics[scale=0.33]{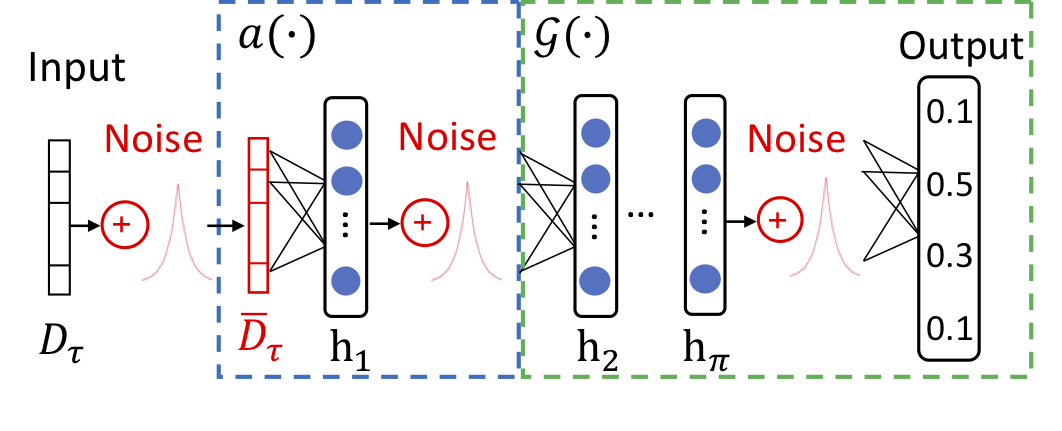}
  \captionof{figure}{Network design of L2DP-ML.} \label{network}
  \includegraphics[scale=0.32]{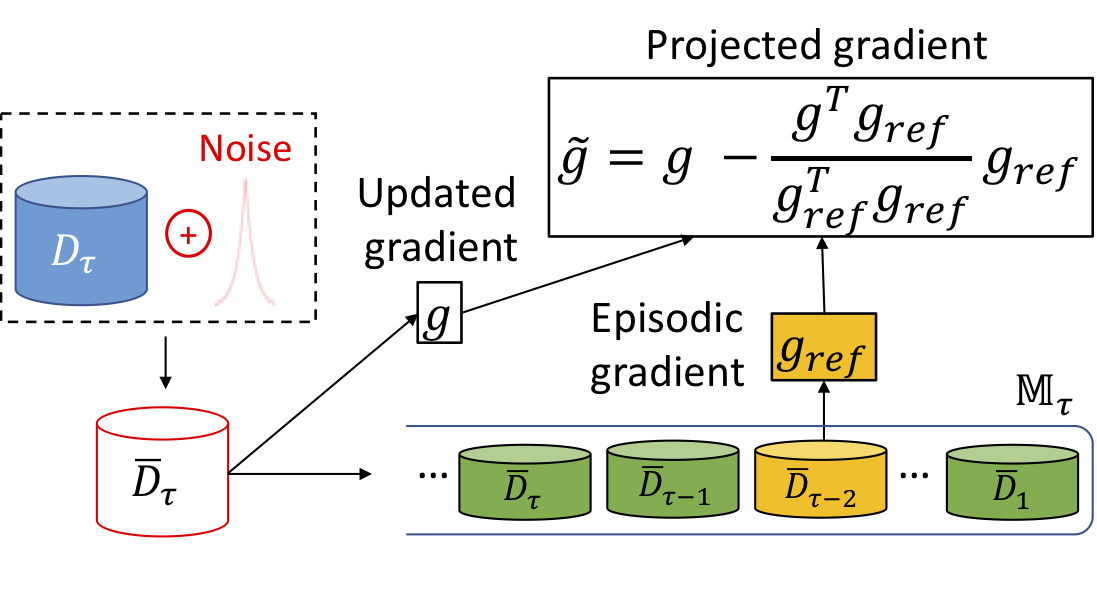}
  \captionof{figure}{Gradient update in L2DP-ML. The updated gradient $\tilde{g}$ is computed by 1) $g_{ref}$ computed from a randomly picked dataset (yellow box) in the episodic memory and 2) $g$ of the current task. }
  \label{grad-update}
\end{minipage}
\end{tabular}

\section{Preserving Lifelong DP}

To overcome the aforementioned issues,
our idea is designing a L2M mechanism such that the privacy budget will not accumulate across training steps while memorizing previously learned tasks. 
More precisely, we design our network as a multi-layer neural network stacked on top of a feature representation learning model. Then, we propose a new Laplace mechanism-based Lifelong DP  algorithm, called \textsc{L2DP-ML} (Alg. \ref{DPL2M-Dataset}), in computing the gradients $g, g_{ref}$, and $\tilde{g}$. Finally, to overcome expensive computation cost and heterogeneity among tasks, we develop a scalable and heterogeneous algorithm through a streaming batch training (Alg. \ref{DPL2M}), to efficiently learn Lifelong DP parameters (Theorem \ref{OverallL2DP}).



\textbf{Network Design.} In our Alg. \ref{DPL2M-Dataset} and Fig.~\ref{network}, a DNN is designed as a stack of an auto-encoder for feature representation learning and a typical multi-layer neural network, as follows: $f(x) = \mathcal{G}(a(x, \theta_1), \theta_2)$ where $a(\cdot)$ is the auto-encoder and $\mathcal{G}(\cdot)$ is the multi-layer neural network. The auto-encoder $a(\cdot)$ takes $x$ as an input with model parameters $\theta_1$; meanwhile, the multi-layer neural network $\mathcal{G}(\cdot)$ takes the output of the auto-encoder $a(\cdot)$ as its input with model parameters $\theta_2$ and returns the class scores $f(x)$.


This network design allows us to: \textbf{(1)} Tighten the sensitivity of our model, since it is easy to train the auto-encoder using less sensitive objective functions, given its small sizes; \textbf{(2)} Reduce the privacy budget consumption, since the computations of the multi-layer neural network is DP when the output of the auto-encoder is DP; and \textbf{(3)} Provide a better re-usability, given that the auto-encoder can be reused and shared for different predictive models.

Given a dataset $D_\tau$, the objective functions of the auto-encoder and the multi-layer neural network can be the classical cross-entropy error functions for data reconstruction at the input layer and for classification at the output layer, denoted $\mathcal{R}_{D_\tau}(\theta_1)$ and $\mathcal{L}_{D_\tau}(\theta_2)$ respectively. Without loss of generality, we define the data reconstruction function
$\mathcal{R}_{D_\tau}(\theta_1)$ and the classification function $\mathcal{L}_{D_\tau}(\theta_2)$ as follows:
\begin{align}
 \mathcal{R}_{D_\tau}(\theta_1)  &= \sum_{x_r \in {D_\tau}} \sum_{s = 1}^{d} \Big[x_{rs}\log (1+e^{-\theta_{1s} h_r}) \Big] + \sum_{x_r \in {D_\tau}} \sum_{s = 1}^{d} \Big[ (1-x_{rs}) \log (1+e^{\theta_{1s} h_r}) \Big] \\
 \mathcal{L}_{D_\tau}(\theta_2)  &= - \sum_{x_r \in {D_\tau}}  \sum_{k=1}^K \Big[ y_{rk} \log (1 + e^{-\mathbf{h}_{\pi r} W^T_{\pi k} }) + (1-y_{rk}) \log (1 + e^{\mathbf{h}_{\pi r} W^T_{\pi k} }) \Big]
\end{align}
where the transformation of  $x_r$ is
$h_r = \theta_1^\top x_r$, the hidden layer $\mathbf{h}_1$ of $a(x, \theta_1)$ given $D_\tau$ is $\mathbf{h}_{1D_\tau} = \{\theta_1^\top x_r\}_{x_r \in D_\tau}$,  $\widetilde{x}_r = \theta_1 h_r$ is the reconstruction of $x_r$, and $\mathbf{h}_{\pi r}$ computed from the $x_r$ through the network with $W_{\pi}$ is the parameter at the last hidden layer $\mathbf{h}_{\pi}$.




Our L2M objective function is defined as:
\begin{align}
& \{\theta_1^{\tau}, \theta_2^{\tau}\} = \arg\min_{\theta_1, \theta_2} [\mathcal{R}_{D_\tau}(\theta_1) + \mathcal{L}_{D_\tau}(\theta_2)]   \label{L2M-DP}  \text{ \ \ \ s.t. \ \ \ } \mathcal{R}_{\mathbb{M}_\tau}(\theta_1^\tau) \leq \mathcal{R}_{\mathbb{M}_\tau}(\theta_{1}^{\tau - 1}) \text{ and } \mathcal{L}_{\mathbb{M}_\tau}(\theta_2^\tau) \leq \mathcal{L}_{\mathbb{M}_\tau}(\theta_{2}^{\tau - 1})  
\end{align} 
where $\{\theta_1, \theta_2\}$ are the model parameters; while, $\{\theta_1^{\tau}, \theta_2^{\tau}\}$ are  the values of $\{\theta_1, \theta_2\}$ after learning task $\tau$.

At each training step on the current task $\tau$, to update the model parameters  $\{\theta_1^{\tau}, \theta_2^{\tau}\}$ minimizing Eq.~\ref{L2M-DP}, we need to compute the gradients $g$ and $g_{ref}$, and then follow Eq.~\ref{agem} to compute the projected gradient $\tilde{g}$ for the model parameters $\{\theta_1^{\tau}, \theta_2^{\tau}\}$ (Fig.~\ref{grad-update}). Given the projected $\tilde{g}$, we can update $\{\theta_1^{\tau}, \theta_2^{\tau}\}$ by applying typical descent operation, as follows.


\paragraph{Gradient Update $g$.}

To compute the gradient $g$ for $\{\theta_1^\tau, \theta_2^\tau\}$ on the current task $\tau$, we first derive polynomial forms of $\mathcal{R}_{D_\tau}(\theta_1)$ and $\mathcal{L}_{D_\tau}(\theta_2)$, by applying the 1st and 2nd orders of Taylor Expansion \citep{tagkey1985} as follows: 
\begin{align}
& \widetilde{\mathcal{R}}_{D_\tau}(\theta_1) \ \  =  \ \  \sum_{x_r \in D_\tau} \sum_{s = 1}^d \Big[ \theta_{1s}\big(\frac{1}{2} - x_{rs}\big)h_r \Big] \\
&\widetilde{\mathcal{L}}_{D_\tau}\big(\theta_2\big) \ \   =  \ \  \sum_{k = 1}^K \sum_{x_r \in D_\tau} \Big[\mathbf{h}_{\pi r} W_{\pi k} - (\mathbf{h}_{\pi r} W_{\pi k}) y_{rk} \Big]    - \sum_{k = 1}^K \sum_{x_r \in D_\tau} \Big[  \frac{1}{2}\lvert \mathbf{h}_{\pi r} W_{\pi k}\lvert  + \frac{1}{8} (\mathbf{h}_{\pi r} W_{\pi k})^2 \Big] 
\end{align}

To preserve $\epsilon_1$-DP in learning $\theta_1$, we leverage Functional Mechanism \citep{zhang2012functional} to inject a Laplace noise 
 into polynomial coefficients of the function $\widetilde{\mathcal{R}}_{D_\tau}(\theta_1)$, which are the input $x$ and the first transformation $\mathbf{h}_1$. Laplace mechanism \citep{Dwork:2014:AFD:2693052.2693053} is well-known in perturbing objective functions to prevent privacy budget accumulation in training ML models \citep{ PhanMLJ2017,NHPhanICDM17,DBLP:journals/corr/abs-1903-09822}. As in \citep{DBLP:journals/corr/abs-1903-09822}, the global sensitivity $\Delta_{\widetilde{\mathcal{R}}}$ is bounded as follows: $\Delta_{\widetilde{\mathcal{R}}} \leq d(\lvert \mathbf{h}_{1}\lvert  + 2)$, with $\lvert \mathbf{h}_{1}\lvert $ is the number of neurons in $\mathbf{h}_{1}$.
The perturbed $\widetilde{\mathcal{R}}$ function becomes:
\begin{equation}
\overline{\mathcal{R}}_{\overline{D}_\tau}(\theta_1) = \sum_{\overline{x}_r \in \overline{D}_\tau} \Big[\sum_{s = 1}^d (\frac{1}{2}\theta_{1s}\overline{h}_r) - \overline{x}_{r} \widetilde{x}_{r} \Big]
\label{PerturbAutoencoder} 
\end{equation} 
where $\overline{x}_{r} = x_{r} + \frac{1}{\lvert \mathcal{D}_\tau\lvert }Lap(\frac{\Delta_{\widetilde{\mathcal{R}}}}{\epsilon_1}), h_r = \theta_1^{\top} \overline{x}_r, \overline{h}_r = h_r + \frac{2}{\lvert \mathcal{D}_\tau\lvert }Lap(\frac{\Delta_{\widetilde{\mathcal{R}}}}{\epsilon_1})$, $\widetilde{x}_{r} = \theta_{1}\overline{h}_r$, $h_r$ is clipped to $[-1, 1]$, and $\epsilon_1$ is a privacy budget.

Importantly, the perturbation of each example $x$ turns the original data $D_\tau$ into a $(\epsilon_1/\gamma_\mathbf{x})$-DP dataset $\overline{D}_\tau = \{\overline{x}_r\}_{x_r \in D_\tau}$ with $\gamma_\mathbf{x}=\Delta_{\widetilde{\mathcal{R}}} / \lvert D_\tau\lvert $ by following Lemma 2 in \citep{DBLP:journals/corr/abs-1903-09822} (Alg. \ref{DPL2M-Dataset}, line 2). 
Based upon that, all the computations on top of the $(\epsilon_1/\gamma_\mathbf{x})$-DP dataset $\overline{D}_\tau$, including $h_r$, $\overline{h}_r$, $\widetilde{x}_r$, and the computation of gradients $g$ of the model parameters $\theta_1$ are shown to be $(\epsilon_1/\gamma_\mathbf{x})$-DP without accessing any additional information from the original data $D_\tau$, i.e., $\forall s \in [1, d]: \nabla_{\theta_{1s}}\overline{\mathcal{R}}_{\overline{D}_\tau}(\theta_{1}) = \frac{\delta \overline{\mathcal{R}}_{\overline{D}_\tau}(\theta_1)}{\delta \theta_{1s}} = \sum_{r = 1}^{\lvert \mathcal{D}_\tau\lvert }\overline{h}_r(\frac{1}{2} - \overline{x}_{rs})$. This follows the post-processing property of DP \citep{Dwork:2014:AFD:2693052.2693053}. Consequently, the total privacy budget used to perturb $\widetilde{\mathcal{R}}$ is $(\epsilon_1 + \epsilon_1/\gamma_{\mathbf{x}})$, by having $\frac{Pr\big(\overline{\mathcal{R}}_{\overline{D}_\tau}(\theta_1)\big)}{Pr\big(\overline{\mathcal{R}}_{\overline{D}'_\tau}(\theta_1)\big)} \times \frac{Pr\big(\overline{D}_\tau \big)}{Pr\big( \overline{D}'_\tau\big)} \leq (\epsilon_1 + \epsilon_1/\gamma_{\mathbf{x}})$. Details are available in our proof of Theorem \ref{OverallL2DP}, Appx.~\ref{ProofT2}.

A similar approach is applied to perturb the objective function $\widetilde{\mathcal{L}}_{D_\tau}(\theta_2)$ at the output layer with a privacy budget $\epsilon_2$. The perturbed function of $\widetilde{\mathcal{L}}$ is denoted as $\overline{\mathcal{L}}_{\overline{D}_\tau}(\theta_{2})$. As in Lemma 3 \citep{DBLP:journals/corr/abs-1903-09822}, the output of  the auto-encoder,  
which is the perturbed transformation $\overline{\mathbf{h}}_{1\overline{D}_\tau} = \{\overline{\theta}_1^{\top} \overline{x}_r + \frac{2}{\lvert \mathcal{D}_\tau\lvert }Lap(\frac{\Delta_{\widetilde{\mathcal{R}}}}{\epsilon_1}) \}_{\overline{x}_r \in \overline{D}_\tau}$, is $(\epsilon_1/\gamma)$-DP, given $\gamma = \frac{2\Delta_{\widetilde{\mathcal{R}}}}{\lvert \mathcal{D}_\tau\lvert  \lVert \overline{\theta}_1 \rVert_{1,1}}$ and $\lVert \overline{\theta}_1 \rVert_{1,1}$ is the maximum 1-norm of $\theta_1$'s columns\footnote{\tiny\url{https://en.wikipedia.org/wiki/Operator\_norm}}.
As a result, the computations of all the hidden layers of the multi-layer neural network $\mathcal{G}(\cdot)$ that takes the output of the auto-encoder $\overline{\mathbf{h}}_{1\overline{D}_\tau}$ as its input, is 
$(\epsilon_1/\gamma)$-DP,  since  $\overline{\mathbf{h}}_{1\overline{D}_\tau}$ is $(\epsilon_1/\gamma)$-DP, following  the post-processing property of DP \citep{Dwork:2014:AFD:2693052.2693053} (Alg. \ref{DPL2M-Dataset}, line 2).

That helps us to \textbf{(1)} avoid extra privacy budget consumption in computing the multi-layer neural network $\mathcal{G}(\cdot)$; \textbf{(2)} tighten the sensitivity of the function $\overline{\mathcal{L}}_{\overline{D}_\tau}$ (i.e., $\Delta_{\mathcal{\widetilde{L}}} \leq 2 \lvert \mathbf{h}_\pi \lvert $); and \textbf{(3)} achieve DP gradient update 
for $\theta_2$. 
The total privacy budget used to perturb $\widetilde{\mathcal{L}}$ is $(\epsilon_1/\gamma + \epsilon_2)$, i.e., $Pr\big(\overline{\mathcal{L}}_{\overline{D}_\tau}(\theta_2)\big) / Pr\big(\overline{\mathcal{L}}_{\overline{D}'_\tau}(\theta_2)\big) \leq (\epsilon_1/\gamma + \epsilon_2)$.
Consequently, the total privacy budget in computing the gradient updates $g$, i.e., $\{\nabla_{\theta_1}\overline{\mathcal{R}}_{\overline{D}_\tau}(\theta^{\tau - 1}_1), \nabla_{\theta_2} \overline{\mathcal{L}}_{\overline{D}_\tau}(\theta^{\tau - 1}_2)\}$, for the current task $\tau$ is $(\epsilon_1 + \epsilon_1/\gamma_{\mathbf{x}} + \epsilon_1/\gamma+\epsilon_2)$-DP (Alg. \ref{DPL2M-Dataset}, lines 5 and 10). \vspace{-5pt}

\paragraph{Episodic and Projected Gradients $g_{ref}$ and $\tilde{g}$.}
Now, we are ready to present our approach in achieving Lifelong DP, by configuring the episodic memory at the current task $\tau$ (i.e., $\mathbb{M}_\tau$) as a \textit{fixed} and \textit{disjoint} set of datasets from previous tasks, i.e.,  $\mathbb{M}_\tau = \{\overline{D}_1, \ldots, \overline{D}_{\tau-1}\}$ (Alg. \ref{DPL2M-Dataset}, line 7); such that, at each training step, the computation of episodic gradients $g_{ref}$ for the model parameters $\{ \theta_1,\theta_2 \}$ using a randomly picked dataset $\overline{D}_{ref} \in \mathbb{M}_\tau$ (Alg. \ref{DPL2M-Dataset}, lines 8 and 11), is $(\epsilon_1 + \epsilon_1/\gamma_{\mathbf{x}} + \epsilon_1/\gamma+\epsilon_2)$-DP, without incurring any additional privacy budget consumption for the dataset $D_{ref}$. The \textit{projected gradients} $\tilde{g}$ is computed from $g$ and $g_{ref}$ (Eq.~\ref{agem}) is also $(\epsilon_1 + \epsilon_1/\gamma_{\mathbf{x}} + \epsilon_1/\gamma+\epsilon_2)$-DP, following the post-processing property of DP \citep{Dwork:2014:AFD:2693052.2693053}. 

Hence, we reformulate the L2M objective function in Eq. \ref{L2M-DP}, as follows:
\begin{align}
& \{\theta^{\tau}_{1}, \theta^{\tau}_{2}\} = \arg\min_{\theta_1, \theta_2} [\overline{\mathcal{R}}_{\overline{D}_\tau}(\theta_1) + \overline{\mathcal{L}}_{\overline{D}_\tau}(\theta_2)] \label{L2M-LifelongDP}  \text{\ \ \ \ s.t. \ \ \ \ } \overline{\mathcal{R}}_{\mathbb{M}_\tau}(\theta^{\tau }_1) \leq \overline{\mathcal{R}}_{\mathbb{M}_\tau}(\theta^{\tau - 1}_{1}), \overline{\mathcal{L}}_{\mathbb{M}_\tau}(\theta^{\tau }_2) \leq \overline{\mathcal{L}}_{\mathbb{M}_\tau}(\theta^{\tau - 1}_{2}) \nonumber \\ 
& \text{where } \mathbb{M}_\tau = \{\overline{D}_1, \ldots, \overline{D}_{\tau-1}\}  
\end{align}

By using the perturbed functions $\overline{\mathcal{R}}$ and $\overline{\mathcal{L}}$, the constrained optimization of Eq. \ref{L2M-LifelongDP} can be addressed similarly to Eq. \ref{agem}, when the projected gradient $\tilde{g}$ is computed as: $\tilde{g} = g - (g^{\top}g_{ref}) / (g^{\top}_{ref} g_{ref}) g_{ref}$, where $g$ is the gradient update on the current task $\tau$, and $g_{ref}$ is computed using a dataset $\overline{D}_{ref}$ randomly selected from the episodic memory $\mathbb{M}_\tau$.

\paragraph{Lifelong DP Guarantee.} Given the aforementioned network
$f(x)$ as the stack of the auto-encoder and the multi-layer neural network, and privacy budgets $\epsilon_1$ and $\epsilon_2$,  the total Lifelong DP privacy consumption in learning the model parameters $\{ \theta_1,\theta_2 \}$ at each task 
is computed in Theorem \ref{OverallL2DP}.

\begin{theorem} Alg. \ref{DPL2M-Dataset} achieves $(\epsilon_1 + \epsilon_1/\gamma_{\mathbf{x}} + \epsilon_1/\gamma+\epsilon_2)$-Lifelong DP in learning $\{\theta^{i}_1, \theta^{i}_2\}_{i \in [1, m]}$.
\label{OverallL2DP}
\end{theorem}

Theorem \ref{OverallL2DP} shows that Alg. \ref{DPL2M-Dataset} achieves $\epsilon$-Lifelong DP in learning the model parameters at each task $\{\theta^{i}_1, \theta^{i}_2\}_{i \in [1, m]}$, where $\epsilon = (\epsilon_1 + \epsilon_1/\gamma_{\mathbf{x}} + \epsilon_1/\gamma+\epsilon_2)$. There are three key properties in the proof of Theorem \ref{OverallL2DP} (Appx.~\ref{ProofT2}): 

\textbf{(1)} For every input $x$ in the whole training set $\overline{\mathcal{D}} = \{\overline{D}_i\}_{i \in [1, m]}$, $x$ is included in \textit{one and only one} dataset, denoted $\overline{D}_x \in \overline{\mathcal{D}}$ (Eq. \ref{PCond2}). Hence, the DP guarantee to $x$ in  $\overline{\mathcal{D}}$ 
is equivalent to the DP guarantee to $x$ in $\overline{D}_x$ (Eqs. \ref{Consistency1} and \ref{LocalDP-Result}).

\textbf{(2)} If we randomly sample tuples from the episodic memory to compute the episodic gradients $g_{ref}$, the sampling set and its neighboring set can have at most $i-1$ different tuples ($i \in [1, m]$), since each $\overline{D}_i$ and its neighboring dataset $\overline{D}'_i$ can have at most 1 different tuple. In addition, a random sampling set of tuples in the episodic memory can overlap with more than one datasets $\overline{D}_i$, which is used to compute the gradient $g$. Importantly, different sampling sets from the episodic memory can overlap each other; thus, a simple data tuple potentially is used in multiple DP-preserving objective functions using these overlapping sets to compute the episodic gradients $g_{ref}$.
These issues introduce additional privacy risk by following the group privacy theory and overlapping datasets in DP. 
We address this problem, by having the episodic memory as a \textit{fixed} and \textit{disjoint} set of datasets across $\mathbb{T}$ training tasks (Eq. \ref{PCond3}). As a result, we can prevent the additional privacy leakage, caused by: \textbf{(i)} Differing at most $i-1$ tuples between neighboring $\mathbb{M}_{i}$ and $\mathbb{M}'_{i}$ for all $i \in (1, m]$; and \textbf{(ii)} Generating new and overlapping sets of data samples for computing the episodic gradient (which are considered overlapping datasets in the parlance of DP) in the typical training. Thus, the optimization on one task does not affect the DP protection of any other tasks, even the objective function given one task can be different from the objective function given other tasks (Eq. \ref{FairDP-Result}).

\textbf{(3)} Together with the results achieved in (1) and (2), by having one and only one privacy budget for every task, we can achieve Eqs. \ref{Cond3} and \ref{Cond4} in Lifelong DP (Def. \ref{LifelongDP}). We present these steps in Eqs. \ref{GlobalDP-Result} and \ref{Consistency-Result}.

\vspace{-5pt}

\section{Scalable and Heterogeneous Training}
\vspace{-5pt}

Although computing the gradients given the whole dataset $\overline{D}_\tau$ achieves Lifelong DP, it has some shortcomings: \textbf{(1)} consumes a large computational memory to store the episodic memory; \textbf{(2)} computational efficiency is low, since we need to use the whole dataset $\overline{D}_\tau$ and $\overline{D}_{ref}$ to compute the gradient update and the episodic gradient at each step; This results in a slow convergence speed and poor utility.

\textbf{Scalability.} To address this, we propose a streaming batch training (Alg. \ref{DPL2M}, Appx. \ref{L2DP-ML with Streaming}), in which a batch of data is used to train the model at each training step, by the following steps. 

\textbf{(1)} Slitting the private training data $\overline{D}_\tau$ ($\forall \tau \in \mathbb{T}$) into disjoint and fixed batches (Alg. \ref{DPL2M}, line 4).

\textbf{(2)} Using a single draw of Laplace noise across batches (Alg. \ref{DPL2M}, lines 1-2). That prevents additional privacy leakage, caused by: (i) Generating multiple draws of noise (i.e., equivalent to applying one DP-preserving mechanism multiple times on the same dataset); (ii) Generating new and overlapping batches (which are considered overlapping datasets in the parlance of DP); and (iii) More importantly, for any example $x$, $x$ is included in \textit{only one} batch. Hence, each \textit{disjoint batch} of data in Alg. \ref{DPL2M} can be considered as a \textit{separate dataset} in Alg. \ref{DPL2M-Dataset}.

\textbf{(3)} For each task, we randomly select a batch to place in the episodic memory (Alg. \ref{DPL2M}, line 17).

\textbf{(4)} At each training step, a batch from the current task is used to compute the gradient $g$, and a batch randomly selected from the episodic memory is used to compute the episodic gradient $g_{ref}$ (Alg. \ref{DPL2M}, lines 11-14). Thus, Alg. \ref{DPL2M} still preserves $(\epsilon_1 + \epsilon_1/\gamma_{\mathbf{x}} + \epsilon_1/\gamma+\epsilon_2)$-Lifelong DP (Theorem \ref{OverallL2DP}). 

By doing so, we significantly reduce the computational complexity and memory consumption, since only a small batch of data from each task is stored in the episodic memory.

\textbf{Heterogeneity.} Based upon this, our algorithm can be applied to address the heterogeneity in terms of data sizes among tasks, which differs from multi-modal tasks  \citep{Liu2019LifelongLF}. We can train one task with multiple epochs, without affecting the Lifelong DP protection in Alg.~\ref{DPL2M}, by 1) keeping all the batches fixed among epochs, and 2) at the end of training each task, we randomly select a batch of that task to place in the episodic memory. The order of the task does not affect the Lifelong DP, since the privacy budget is not accumulated across tasks. These distinct properties enable us to customize our training, by having different numbers of training epochs for different tasks and having different training orders of tasks. Tasks with \textit{smaller numbers of data tuples} can have \textit{larger numbers of training epochs.} This helps us to achieve better model utility under the same privacy protection as shown in our experiments.





\vspace{-5pt}
\section{Experiments} 
\vspace{-5pt}

Our validation focuses on understanding the impacts of the privacy budget $\epsilon$ and the heterogeneity on model utility. For reproducibility, our implementation 
is available and uploaded.


\textbf{Baseline Approaches.}
We consider \textbf{A-gem} \citep{chaudhry2018efficient}  as an  upper bound in terms of model performance, since A-gem is a noiseless model. 
We aim to show how much model utility is compromised for the Lifelong DP protection. Also, we consider the naive algorithm \citep{desai2021continual}, called \textbf{NaiveGaussian}, as a baseline to demonstrate the effectiveness of our \textsc{L2DP-ML} mechanism. It is worth noting that there is a lack of a precise definition of adjacent databases resulting in an unclear or not well-justified DP protection for L2M in existing works \citep{farquhar2019differentially, phan2019private}. Therefore, we do not consider them as baselines in our experiments.

To evaluate the heterogeneity, we further derive several versions of our algorithm (Alg.~\ref{DPL2M}), including: \textbf{(1)} \textbf{Balanced \textsc{L2DP-ML}}, in which all the tasks have the same number of training steps, given a fixed batch size. This is also true for a \textbf{Balanced A-gem} algorithm; \textbf{(2)} \textbf{L2DP-ML} with the same number of epochs for all the tasks; and \textbf{(3)} \textbf{Heterogeneous L2DP-ML}, in which a fixed number of training epochs is assigned to each task. The numbers of epochs among tasks can be different. For instance, 5 epochs are used to train tasks with 5Hz, 10Hz, and 20Hz data, and 1 epoch is used to train the task with a larger volume of 50Hz data. The number of epochs is empirically identified by the data size of each task, since the search space of the number of epochs for each task is exponentially large. 



\textbf{Datasets.} We evaluate our approach using permuted and split MNIST \citep{Kirkpatrick3521}, permuted and split CIFAR-10 \citep{ivan2019convolutional}, split CIFAR-100 datasets\footnote{Datasets were downloaded and evaluated by Phung Lai, Han Hu, and NhatHai Phan.}, and our human activity recognition in the wild (HARW) dataset. Permuted MNIST is a variant of MNIST \citep{726791} dataset, where each task has a random permutation of the input pixels, which is applied to all the images of that task. We adopt this approach to permute the CIFAR-10 dataset, including the input pixels and three color channels. 
Our HARW dataset was collected from 116 users, each of whom provided mobile sensor data and labels for their activities on Android phones consecutively in three months. HARW is an ultimate task for L2M, since different sensor sampling rates, e.g., 50Hz, 20Hz, 10Hz, and 5Hz, from different mobile devices are considered as L2M tasks. The classification output includes five classes of human activities, i.e., walking, sitting, in car, cycling, and running. The data collection and processing of our HARW dataset is in Appx.~\ref{HARWData}. The setting of split CIFAR-10 and CIFAR-100, and split MNIST datasets are in Appx. \ref{Hyper-parameter Search}.


 \begin{figure*}[t]
  \centering
\subfloat[]{\label{FigMNISTCIFARa}\includegraphics[scale=0.15]{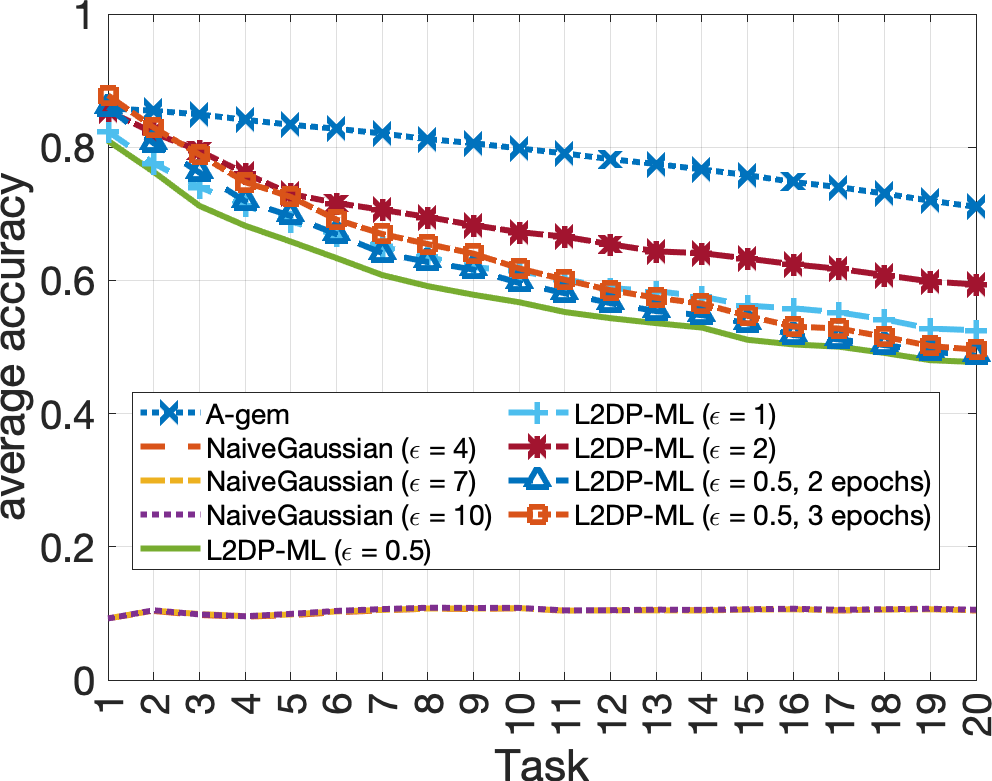}}\hspace*{0.2cm}
\subfloat[]{\label{FigMNISTCIFARb}\includegraphics[scale=0.15]{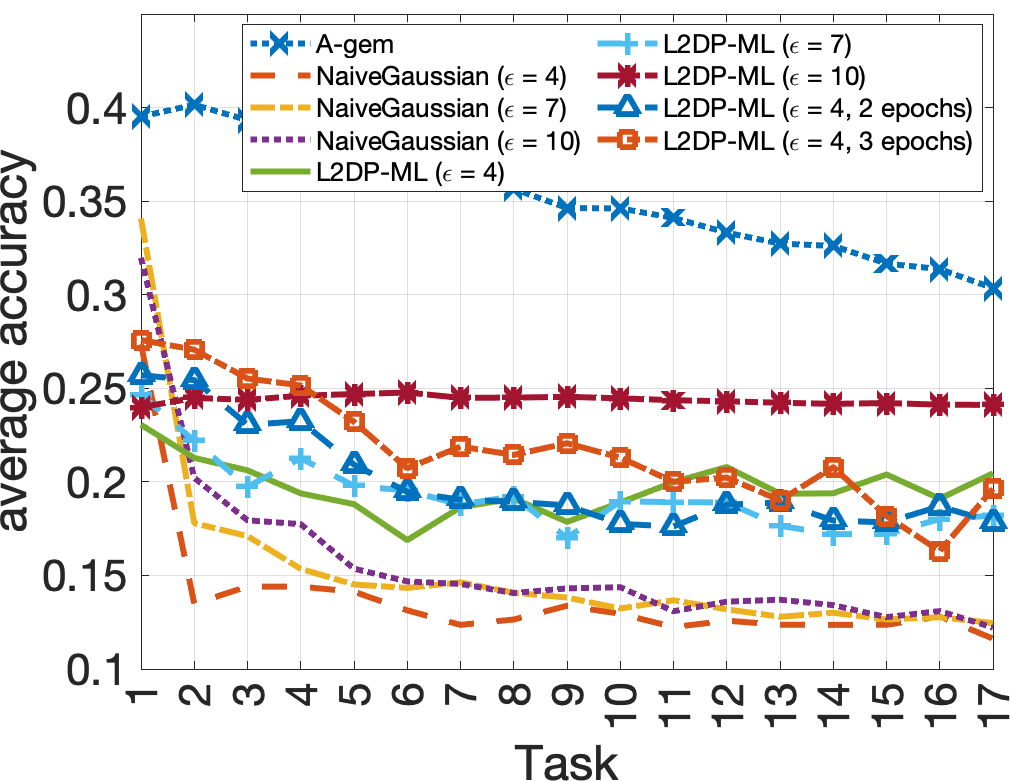}}
\hspace*{0.2cm}
\subfloat[]{\label{FigMNISTCIFARc}\includegraphics[scale=0.15]{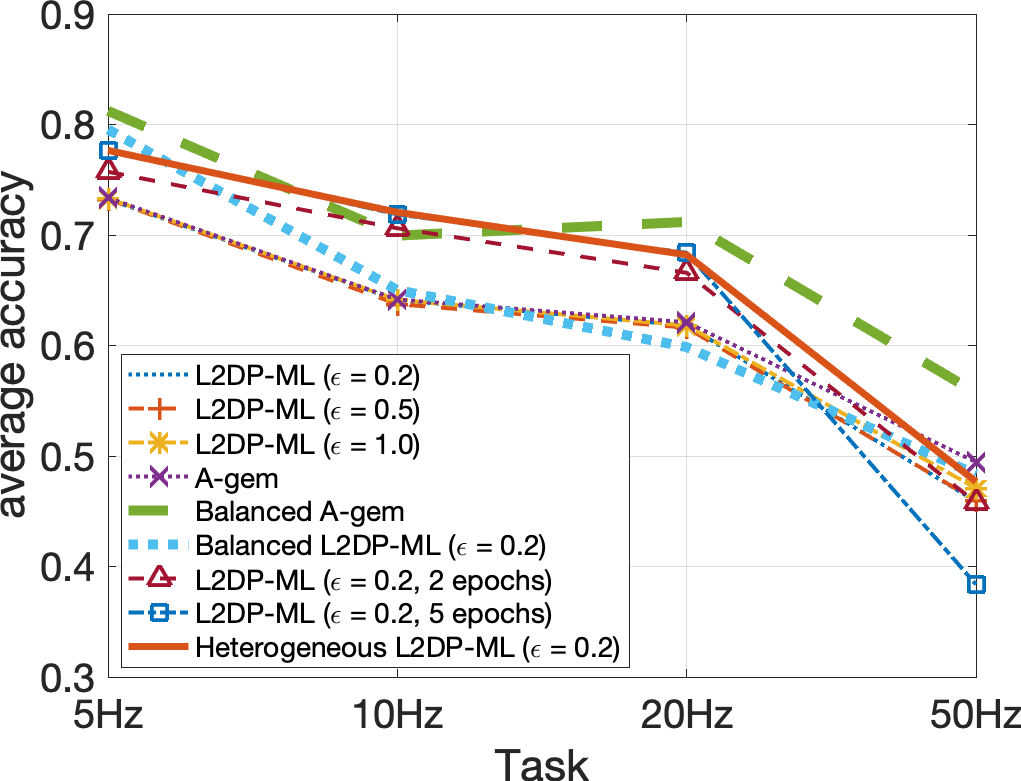}} \vspace{-10pt} 
\caption{Average accuracy in the (a) Permuted MNIST (20 tasks), b) Permuted CIFAR-10 (17 tasks), and (c) HARW.} \vspace{-10pt} 
\label{FigMNISTCIFAR} 
 \end{figure*}

\begin{table*}[t]
\centering
\caption{Average forgetting measure. (Smaller the better) } \vspace{-10pt}
\label{forgetting}
\small
\begin{tabular}{| c | c| c | c| c | c| c | c| c | c|}
\hline
\multicolumn{2}{|l|}{} & \textsc{L2DP-ML} & NaiveGaussian  & A-gem \\
\hline
\multirow{3}*{\makecell{Permuted MNIST}}  &  $\epsilon = 0.5$ &   0.305 $\pm$ 0.00886 & 0.012 $\pm$ 0.00271 & \multirow{3}*{0.162 $\pm$ 0.01096}\\
\cline{2-4}
&  $\epsilon = 1$ & 0.278 $\pm$ 0.00907  &  0.015 $\pm$ 0.00457  &   \\
\cline{2-4}
&  $\epsilon = 2$ &   0.237 $\pm$ 0.00586 &0.017 $\pm$ 0.00385   &  \\
\hline  
\multirow{3}*{\makecell{Permuted CIFAR-10}}  &  $\epsilon = 4$ &   0.033 $\pm$ 0.00896 & 0.138 $\pm$ 0.00582 & \multirow{3}*{0.133 $\pm$ 0.00859}\\
\cline{2-4}
&  $\epsilon = 7$ &  0.062 $\pm$ 0.01508 &  0.174 $\pm$ 0.01149  & \\
\cline{2-4}
&  $\epsilon = 10$ &    0.034 $\pm$ 0.00184 & 0.181 $\pm$ 0.01956   & \\
\hline 
\multicolumn{2}{|l|}{}  & L2DP-ML & Balanced L2DP-ML ($\epsilon = 0.2$) &  A-gem  \\
\hline  
\multirow{5}*{\makecell{HARW \\ (5Hz - 10Hz \\- 20Hz - 50Hz) }}  &  $\epsilon = 0.2$ & 0.1133 $\pm$ 0.0003 & \multirow{3}*{0.1309 $\pm$ 0.002}  & \multirow{3}*{0.1269 $\pm$ 0.00045 } \\
\cline{2-3}
& $\epsilon = 0.2$ (2 epochs) & 0.1639 $\pm$ 0.00074 &    &  \\
\cline{2-3}
& $\epsilon = 0.2$ (5 epochs) & 0.2031 $\pm$ 0.0013 &   &  \\
\cline{2-5}
&  $\epsilon = 0.5$ & 0.1124 $\pm$ 0.00029 & Heterogeneous \textsc{L2DP-ML} ($\epsilon = 0.2$) &  Balanced A-gem  \\
\cline{2-5}
&  $\epsilon = 1$ & 0.1106 $\pm$ 0.00026  & 0.1920 $\pm$ 0.00034 &  0.1593 $\pm$ 0.00021 \\
\hline 
\end{tabular} \par \vspace{-20pt}
\end{table*}  

\textbf{Model Configuration.} In the permuted MNIST dataset, we used three convolutional layers (32, 64, and 96 features). 
In the permuted CIFAR-10 dataset, we used a Resnet-18 network (64, 64, 128, 128, and 160 features) with kernels (4, 3, 3, 3, and 3). 
In the HARW dataset, we used three convolutional layers (32, 64, and 96 features).  
Detailed model configurations are in the Appx. \ref{Hyper-parameter Search}. To conduct a fair comparison, we applied a \textit{grid-search} for the best values of hyper-parameters, including the privacy budget $\epsilon \in [4, 10]$, the noise scale $z \in [1.1, 2.5]$, and the clipping bound $C \in [0.01, 1]$, in the NaiveGaussian mechanism. Based on the results of our hyper-parameter grid-search (Table \ref{tab:cifar-10}), we set $z=2.2$ for $\epsilon=4.0$, $z=1.7$ for $\epsilon=7.0$, and $z=1.4$ for $\epsilon=10.0$, and $C=0.01$ is used for all values of $\epsilon$.

\textbf{Evaluation Metrics.} We employ the well-applied average accuracy and forgetting measures after the model has been trained with all the batches up to task $\tau$ \citep{DBLP:journals/corr/abs-1801-10112,chaudhry2018efficient}, defined as follows: \textbf{(1)}
$
\textit{average accuracy}_\tau = \frac{1}{\tau}\sum_{t=1}^{\tau} a_{\tau, n, t}
$,
where $a_{\tau, n, t} \in [0, 1]$ is the accuracy evaluated on the test set of task $t$, after the model has been trained with the $n^{th}$ batch of task $\tau$, and the training dataset of each task, $D_\tau$, consists of a total $n$ batches; \textbf{(2)}
$
\textit{average forgetting}_\tau = \frac{1}{\tau - 1} \sum_{t = 1}^{\tau - 1} f^\tau_t
$,
where $f^\tau_t$ is the forgetting on task $t$ after the model is trained with all the batches up till task $\tau$. $f^\tau_t$ is computed as follows: $f^\tau_t = \max_{l \in \{1, \ldots, \tau - 1\}} (a_{l, n, t} - a_{\tau, n, t})$; and \textbf{(3)} We measure the significant difference between two average accuracy curves induced by  models $A$ and $B$ after task $\tau$, using a $p$ value (2-tail t-tests) curve:
$p\text{\ }value = \big(\{\frac{1}{i}\sum_{t = 1}^i a^{(A)}_{i, n, t}\}_{i \in [1, \tau]}, \{\frac{1}{i}\sum_{t = 1}^i a^{(B)}_{i, n, t}\}_{i \in [1, \tau]}\big)$.
All statistical tests are 2-tail t-tests.

 \begin{figure*}[t]
  \centering
\subfloat[]{\label{HARExtraa}\includegraphics[scale=0.15]{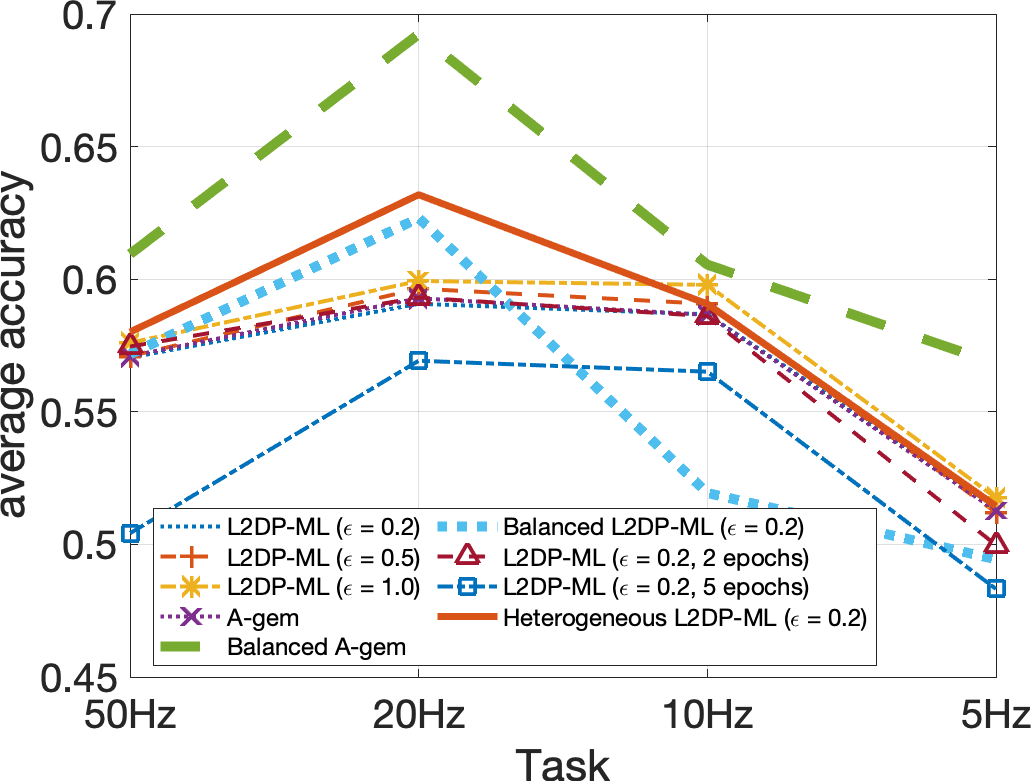}}\hspace*{0.1cm}
\subfloat[]{\label{HARExtrab}\includegraphics[scale=0.15]{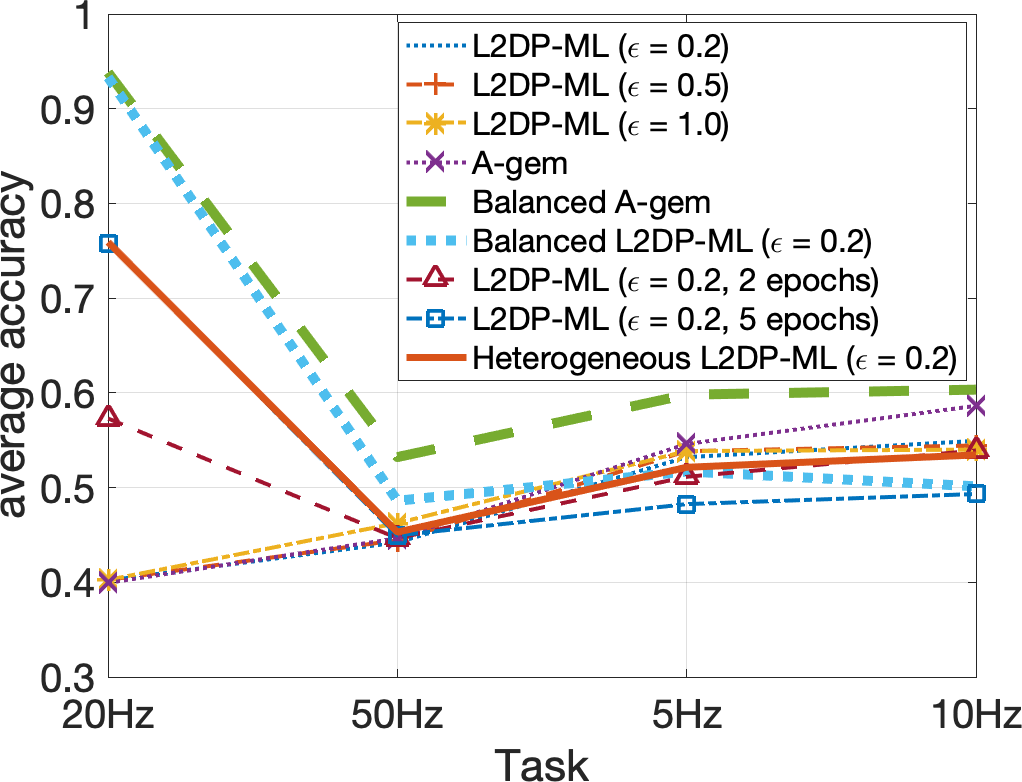}}
\hspace*{0.1cm}
\subfloat[]{\label{HARExtrac}\includegraphics[scale=0.15]{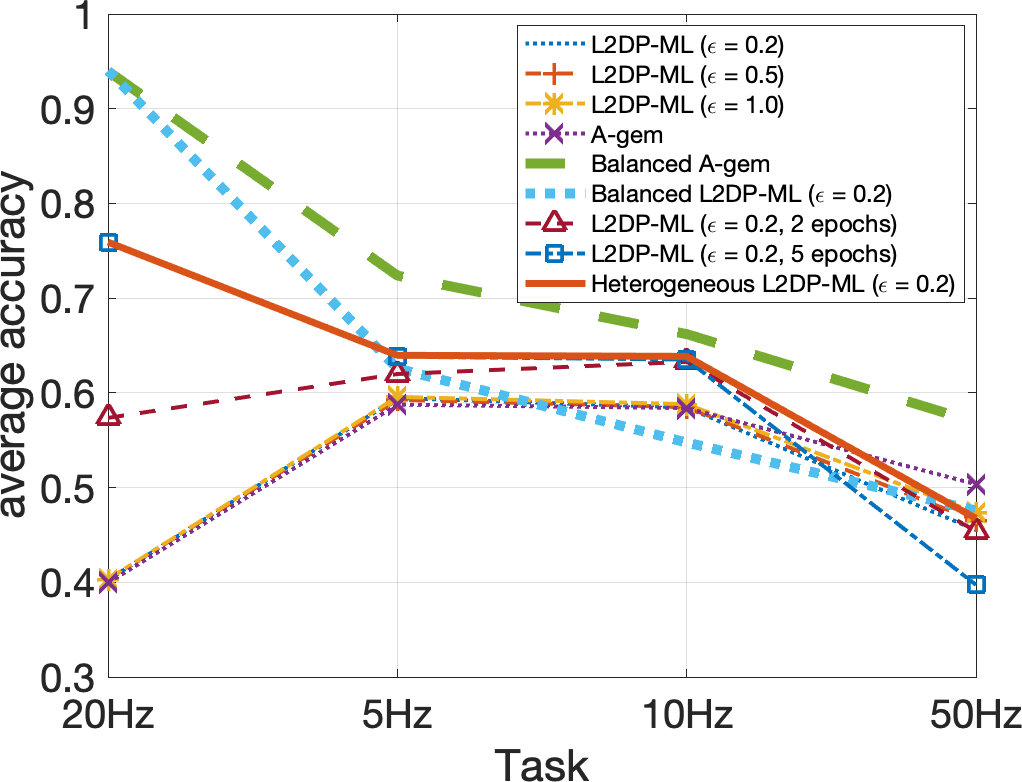}} \vspace{-10pt}
\caption{Average accuracy in the HARW dataset with random task orders: (a) HARW 50Hz - 20Hz - 10Hz - 5Hz, (b) HARW 20Hz - 50Hz - 5Hz - 10Hz, and (c) HARW 20Hz - 5Hz - 10Hz - 50Hz (higher the better).} \vspace{-10pt}
\label{HARExtra}  
 \end{figure*}

\begin{table*}[t]
\caption{Average forgetting measure on random orders of HARW tasks. The order of [20Hz, 5Hz, 10Hz, 50Hz] is in Table \ref{SuppHARTask}, Appx. \ref{Hyper-parameter Search}. (Smaller the better)} \vspace{-13pt}
\label{Extraforgetting}
\begin{center}
\resizebox{\textwidth}{!}{%
\begin{tabular}{| c | c|c|c|}
\hline 
 & L2DP-ML ($\epsilon = 0.2$) & L2DP-ML ($\epsilon = 0.5$) & L2DP-ML ($\epsilon = 1$) \\
 \cline{2-4}
 & 0.1016 $\pm$ 0.0002 & 0.1012 $\pm$ 0.0001 & 0.098 $\pm$ 0.0001 \\ 
 \cline{2-4}
HARW (50Hz - & A-gem & Balanced A-gem & Balanced L2DP-ML ($\epsilon = 0.2$) \\
 \cline{2-4}
20Hz - 10Hz - 5Hz) & 0.1029 $\pm$ 0.0002 & 0.1241 $\pm$ 0.0002 & 0.1274 $\pm$ 0.0008 \\  \cline{2-4}
 & L2DP-ML ($\epsilon = 0.2$, 2 epochs) & L2DP-ML ($\epsilon = 0.2$, 5 epochs) & Heterogeneous L2DP-ML ($\epsilon = 0.2$) \\ \cline{2-4}
 & 0.1148 $\pm$ 0.0002 & 0.1012 $\pm$ 0.0014 & 0.1442 $\pm$ 0.0003 \\
 \hline
\end{tabular}} \par \vspace{-5pt}
\end{center}

\begin{center}
\resizebox{\textwidth}{!}{%
\begin{tabular}{| c | c|c|c|}
\hline 
 & L2DP-ML ($\epsilon = 0.2$) & L2DP-ML ($\epsilon = 0.5$) & L2DP-ML ($\epsilon = 1$) \\
 \cline{2-4}
 & 0.0769 $\pm$ 2.07e-5 & 0.0761 $\pm$ 3.88e-5 & 0.0772 $\pm$ 6.7e-5 \\ 
 \cline{2-4}
HARW (20Hz - & A-gem & Balanced A-gem & Balanced L2DP-ML ($\epsilon = 0.2$) \\
 \cline{2-4}
50Hz - 5Hz - 10Hz) & 0.0781 $\pm$ 2.28e-5 & 0.14 $\pm$ 3.26e-4 & 0.1248 $\pm$ 0.0013 \\  \cline{2-4}
 & L2DP-ML ($\epsilon = 0.2$, 2 epochs) & L2DP-ML ($\epsilon = 0.2$, 5 epochs) & Heterogeneous L2DP-ML ($\epsilon = 0.2$) \\ \cline{2-4}
 & 0.0775 $\pm$ 8.45e-5 & 0.099 $\pm$ 0.0015 & 0.1268 $\pm$ 0.00028 \\
 \hline
\end{tabular}} \par \vspace{-21pt}
\end{center}

\end{table*}


\textbf{Results in Permuted MNIST.}  Fig.~\ref{FigMNISTCIFARa} and Table \ref{forgetting} illustrate the average accuracy and forgetting measure of each model as a function of the privacy budget $\epsilon$ on the permuted MNIST dataset. It is clear that the NaiveGaussian mechanism does not work well under a tight privacy budget $\epsilon \in [0.5, 2]$ given a large number of tasks $m = 20$. This is because each task can consume a tiny privacy budget $\epsilon/m$ resulting in either a large noise injected into the clipped gradients or a lack of training steps to achieve better model utility. By avoiding the privacy budget accumulation across tasks and training steps, our L2DP-ML models significantly outperform the NaiveGaussian mechanism. Our L2DP-ML model achieves 47.73\% compared with 10.43\% of the NaiveGaussian after 20 tasks given $\epsilon = 0.5$ ($p < 6.81e-15$). 

Regarding the upper bound performance, there is a small average accuracy gap  between the noiseless A-gem model and our L2DP-ML models given a small number of tasks. The gap increases when the number of tasks increases (23.3\% at $\epsilon = 0.5$ with 20 tasks). The larger the privacy budget (i.e., $\epsilon = 2.0$), the higher the average accuracy we can achieve, i.e., an improvement of 9.92\% with $p < 2.83e-14$, compared with smaller privacy budgets (i.e., $\epsilon = 0.5$). 
Also, our L2DP-ML models have a relatively good average forgetting with tight privacy protection ($\epsilon = 0.5, 1,$ and $2$), compared with the noiseless A-gem model. 

\textbf{Results in Permuted CIFAR-10.}
Although permuted CIFAR-10 tasks are very difficult (Fig.~\ref{FigMNISTCIFARb} and Table \ref{forgetting}), even with the noiseless A-gem model, i.e., 35.24\% accuracy on average, the results on the permuted CIFAR-10 further strengthen our observation. Our L2DP-ML models significantly outperform the NaiveGaussian mechanism. Our L2DP-ML model achieves an improvement of 8.84\% in terms of average accuracy over the NaiveGaussian after 17 tasks given $\epsilon = 4$ ($p < 4.68e-7$). We further observe that the NaiveGaussian mechanism has a remarkably larger average forgetting compared with our L2DP-ML  (Table \ref{forgetting}).

Interestingly, 
the gap between A-gem and our L2DP-ML models is notably shrunken when the number of tasks increases (from 16.47\% with 1 task to 9.89\% with 17 tasks, at $\epsilon = 4$). In addition, the average forgetting values in our L2DP-ML are better than the noiseless A-gem. This is a promising result.
We also registered that the larger the privacy budget (i.e., $\epsilon = 10$), the higher the average accuracy that we can achieve, i.e., an improvement of 4.73\% with $p < 1.15e-9$, compared with smaller budgets (i.e., $\epsilon = 4$).
 

\textbf{Heterogeneous Training.} 
We now focus on shedding light into understanding the impacts of heterogeneity and privacy on model utility given different variants of our L2DP-ML mechanisms and the noiseless A-gem model. The $p$ value curves are in Figures \ref{PvalueMNISTCIFAR} and \ref{HARExtraPValue}, Appx.~\ref{Hyper-parameter Search}.

On the HARW task (Fig.~\ref{FigMNISTCIFARc} and Table \ref{forgetting}), our L2DP-ML model achieves a very competitive average accuracy, given a very tight DP budget $\epsilon = 0.2$ (i.e., 61.26\%) compared with the noiseless A-gem model (i.e., 62.27\%), across four tasks. Our model also achieves a better average forgetting, i.e., 11.33, compared with 12.69 of the noiseless A-gem model. That is promising. Increasing the privacy budget modestly increases the model performance. The differences in terms of average accuracy and forgetting are not significant. This is also true, when we randomly flip the order of the tasks (Fig.~\ref{HARExtra} and Table \ref{Extraforgetting}). The results showed that our model  effectively preserves Lifelong DP in HARW tasks.
 

Heterogeneous training, with customized numbers of epochs and task orders, further improves our model performance, under the same Lifelong DP protection. Fig.~\ref{PvalueMNISTCIFAR}  illustrates the $p$ values between the average accuracy curves of our L2DP-ML, given \textbf{1)} heterogeneous training with different numbers of epochs, \textbf{2)} task orders, and \textbf{3)} privacy budgets, over its basic settings, i.e., $\epsilon = 0.5$ for the permuted MNIST dataset, $\epsilon = 4$ for the permuted CIFAR-10 dataset, and $\epsilon = 0.2$ for the HARW dataset, with one training epoch. 

$\bullet$ In the permuted MNIST dataset (Figs.~\ref{FigMNISTCIFARa} and \ref{PvalueMNISTCIFARa}), when our L2DP-ML model is trained with 2 or 3 epochs per task, 
the average accuracy is improved, i.e., 2.81\%, 4.8\% given 2, 3 epochs, respectively, with $p < 8.44 e-9$. In the permuted CIFAR-10, using larger numbers of training epochs shows significant performance improvements over a small number of tasks (Fig.~\ref{PvalueMNISTCIFARb}). When the number of tasks becomes larger, the $p$ values become less significant (even insignificant), compared with the $p$ value curves of larger DP budgets (i.e., $\epsilon = 2$ and $\epsilon = 10$ in the permuted MNIST and permuted CIFAR-10). Meanwhile, training with a larger number of epochs yields better results with small numbers of tasks (i.e., fewer than 6 tasks), compared with larger DP budgets. 

$\bullet$ In the HARW tasks, the improvement is more significant (Figs.~\ref{FigMNISTCIFARc} and \ref{PvalueMNISTCIFARc}). Heterogeneous and Balanced L2DP-ML models outperform the basic settings with uniform numbers of training epochs, i.e., 1, 2, and 5 epochs. On average, we registered an improvement of 1.93\% given the Balanced L2DP-ML and an improvement of 5.14\% given the Heterogeneous L2DP-ML, over the basic setting (1 training epoch). The results are statistically significant (Fig.~\ref{PvalueMNISTCIFARc}). The average forgetting values of the Balanced L2DP-ML (0.1593) and the Heterogeneous L2DP-ML (0.1920) are higher than the basic setting (0.1133), with $p < 2.19e-5$ (Table \ref{forgetting}). This is expected as a primary trade-off in L2M, given a better average accuracy. In fact, the average forgetting values are also notably higher given larger uniform numbers of epochs, i.e, 2 and 5 epochs, and the Balanced A-gem. We do not address this fundamental issue in L2M since it is out-of-scope of this study. We focus on preserving Lifelong DP. 

$\bullet$ We observe similar results in randomly flipping the order of the tasks (Figs.~\ref{HARExtra} and \ref{HARExtraPValue},  Table \ref{Extraforgetting}). Among all task orders, our Heterogeneous L2DP-ML achieves the best average accuracy (66.4\%) with the task order [5Hz, 10Hz, 20Hz, 50Hz] (Fig.~\ref{FigMNISTCIFARc}) compared with the worse order [20Hz, 50Hz, 5Hz, 10Hz] (56.69\%) (Fig.~\ref{HARExtrab}), i.e., $p < 9.9e-5$. 
More importantly, in both average accuracy and forgetting, our Balanced and Heterogeneous L2DP-ML models achieve a competitive performance compared with the noiseless Balanced A-gem, which is considered to have the upper bound performance, and a better performance compared with having the uniform numbers of epochs across tasks. This obviously shown that the distinct ability to offer the heterogeneity in training across tasks greatly improves our model performance, under the same Lifelong DP protection.

\textbf{Results in Split Tasks.} We observe similar results on split CIFAR-10, CIFAR-100, and MNIST datasets as L2DP-ML achieves competitive average accuracy approaching the noiseless A-gem model under rigorous privacy budgets (Fig. \ref{FigMNISTCIFARCoLLas}, Appx. \ref{Hyper-parameter Search}). After 5 tasks of the split MNIST dataset, L2DP-ML achieves 73.54\% and 81.83\% in average accuracy at the privacy budgets $0.5$ and $1$ respectively, compared with 79.71\% of the noiseless A-gem. Interestingly, our L2DP-ML has slightly higher average accuracy than the noiseless A-gem after 11 tasks of the split CIFAR-10 and CIFAR-100 dataset (14.83\% in L2DP-ML at $\epsilon = 4$ compared with 13.44\% in the noiseless A-gem). One reason is that Lifelong DP-preserving noise can help to mitigate the catastrophic forgetting. As showed in Table \ref{forgetting-appx} (Appx. \ref{Hyper-parameter Search}), our L2DP-ML obtains a significantly lower average forgetting (2.7\% at $\epsilon = 4$) than the noiseless A-gem (19.5\%).

\vspace{-7.5pt}
\section{Conclusion}
\label{con}
\vspace{-7.5pt}

In this paper, we showed that L2M introduces unknown privacy risk and challenges in preserving DP. To address this, we established a connection between DP preservation and L2M, through a new definition of Lifelong DP. 
To preserve Lifelong DP, we proposed the first scalable and heterogeneous mechanism, called L2DP-ML. Our model shows promising results in several tasks with different settings and opens a long-term avenue to achieve better model utility with lower computational cost, under Lifelong DP.

\section*{Acknowledgement}

This work is partially supported by grants NSF IIS-2041096 / 2041065, NSF CNS-1935928 / 1935923, NSF CNS-1850094, and Qualcomm Incorporated.



\bibliography{collas2022_conference}

\begin{thebibliography}{49}
\providecommand{\natexlab}[1]{#1}
\providecommand{\url}[1]{\texttt{#1}}
\expandafter\ifx\csname urlstyle\endcsname\relax
  \providecommand{\doi}[1]{doi: #1}\else
  \providecommand{\doi}{doi: \begingroup \urlstyle{rm}\Url}\fi

\bibitem[Abadi et~al.(2016)Abadi, Chu, Goodfellow, McMahan, Mironov, Talwar,
  and Zhang]{Abadi}
M.~Abadi, A.~Chu, I.~Goodfellow, H.B. McMahan, I.~Mironov, K.~Talwar, and
  L.~Zhang.
\newblock Deep learning with differential privacy.
\newblock In \emph{{ACM SIGSAC} CCS}, pp.\  308--318, 2016.

\bibitem[Abadi et~al.(2017)Abadi, Erlingsson, Goodfellow, McMahan, Mironov,
  Papernot, Talwar, and Zhang]{abadi2017protection}
M.~Abadi, U.~Erlingsson, I.~Goodfellow, H.B. McMahan, I.~Mironov, N.~Papernot,
  K.~Talwar, and L.~Zhang.
\newblock On the protection of private information in machine learning systems:
  Two recent approches.
\newblock In \emph{IEEE CSF}, pp.\  1--6, 2017.

\bibitem[Abati et~al.(2020)Abati, Tomczak, Blankevoort, Calderara, Cucchiara,
  and Bejnordi]{abati2020conditional}
D.~Abati, J.~Tomczak, T.~Blankevoort, S.~Calderara, R.~Cucchiara, and B.E.
  Bejnordi.
\newblock Conditional channel gated networks for task-aware continual learning.
\newblock In \emph{CVPR}, pp.\  3931--3940, 2020.

\bibitem[Arfken(1985)]{tagkey1985}
G.B. Arfken.
\newblock Mathematical methods for physicists (third edition), 1985.

\bibitem[Chaudhry et~al.(2018)Chaudhry, Dokania, Ajanthan, and
  Torr]{DBLP:journals/corr/abs-1801-10112}
A.~Chaudhry, P.K. Dokania, T.~Ajanthan, and P.H.S. Torr.
\newblock Riemannian walk for incremental learning: Understanding forgetting
  and intransigence.
\newblock In \emph{ECCV}, pp.\  532--547, 2018.

\bibitem[Chaudhry et~al.(2019)Chaudhry, Ranzato, Rohrbach, and
  Elhoseiny]{chaudhry2018efficient}
A.~Chaudhry, M.~Ranzato, M.~Rohrbach, and M.~Elhoseiny.
\newblock Efficient lifelong learning with a-gem.
\newblock \emph{ICLR}, 2019.

\bibitem[Choi et~al.(2017)Choi, Schuetz, Stewart, and Sun]{Choiocw112}
E.~Choi, A.~Schuetz, W.F. Stewart, and J.~Sun.
\newblock Using recurrent neural network models for early detection of heart
  failure onset.
\newblock \emph{Journal of the American Medical Informatics Association},
  24\penalty0 (2):\penalty0 361--370, 2017.

\bibitem[Desai et~al.(2021)Desai, Lai, Phan, and Thai]{desai2021continual}
P.~Desai, P.~Lai, N.H. Phan, and M.T. Thai.
\newblock Continual learning with differential privacy.
\newblock In \emph{International Conference on Neural Information Processing},
  pp.\  334--343, 2021.

\bibitem[Dwork \& Lei(2009)Dwork and Lei]{10.1145/1536414.1536466}
C.~Dwork and J.~Lei.
\newblock Differential privacy and robust statistics.
\newblock In \emph{ACM STOC}, pp.\  371--380, 2009.

\bibitem[Dwork et~al.(2006)Dwork, McSherry, Nissim, and
  Smith]{dwork2006calibrating}
C.~Dwork, F.~McSherry, K.~Nissim, and A.~Smith.
\newblock Calibrating noise to sensitivity in private data analysis.
\newblock In \emph{Theory of cryptography conference}, pp.\  265--284, 2006.

\bibitem[Dwork et~al.(2014)Dwork, Roth, et~al.]{Dwork:2014:AFD:2693052.2693053}
C.~Dwork, A.~Roth, et~al.
\newblock The algorithmic foundations of differential privacy.
\newblock \emph{Foundations and Trends in Theoretical Computer Science},
  9\penalty0 (3-4):\penalty0 211--407, 2014.

\bibitem[Ebrahimi et~al.(2020)Ebrahimi, Meier, Calandra, Darrell, and
  Rohrbach]{ebrahimi2020adversarial}
S.~Ebrahimi, F.~Meier, R.~Calandra, T.~Darrell, and M.~Rohrbach.
\newblock Adversarial continual learning.
\newblock In \emph{ECCV}, pp.\  386--402, 2020.

\bibitem[Farquhar \& Gal(2018)Farquhar and Gal]{farquhar2019differentially}
S.~Farquhar and Y.~Gal.
\newblock Differentially private continual learning.
\newblock \emph{Privacy in Machine Learning and AI workshop at ICML}, 2018.

\bibitem[Fredrikson et~al.(2015)Fredrikson, Jha, and
  Ristenpart]{Fredrikson:2015:MIA}
M.~Fredrikson, S.~Jha, and T.~Ristenpart.
\newblock Model inversion attacks that exploit confidence information and basic
  countermeasures.
\newblock In \emph{ACM SIGSAC Conference on Computer and Communications
  Security}, pp.\  1322--1333, 2015.

\bibitem[He \& Zhu(2022)He and Zhu]{he2022online}
J.~He and F.~Zhu.
\newblock Online continual learning via candidates voting.
\newblock In \emph{Proceedings of the IEEE/CVF Winter Conference on
  Applications of Computer Vision}, pp.\  3154--3163, 2022.

\bibitem[Helmstaedter et~al.(2013)Helmstaedter, Briggman, Turaga, Jain, Seung,
  and Denk]{helmstaedter2013}
M.~Helmstaedter, K.L. Briggman, S.C. Turaga, V.~Jain, H.S. Seung, and W.~Denk.
\newblock Connectomic reconstruction of the inner plexiform layer in the mouse
  retina.
\newblock \emph{Nature}, 500\penalty0 (7461):\penalty0 168--174, 2013.

\bibitem[Ignatov(2018)]{ignatov2018real}
A.~Ignatov.
\newblock Real-time human activity recognition from accelerometer data using
  convolutional neural networks.
\newblock \emph{Applied Soft Computing}, 62:\penalty0 915--922, 2018.

\bibitem[Ivan(2019)]{ivan2019convolutional}
C.~Ivan.
\newblock Convolutional neural networks on randomized data.
\newblock In \emph{CVPR Workshops}, pp.\  1--8, 2019.

\bibitem[Kirkpatrick et~al.(2017)Kirkpatrick, Pascanu, Rabinowitz, Veness,
  Desjardins, Rusu, Milan, Quan, Ramalho, Grabska-Barwinska,
  et~al.]{Kirkpatrick3521}
J.~Kirkpatrick, R.~Pascanu, N.~Rabinowitz, J.~Veness, G.~Desjardins, A.A. Rusu,
  K.~Milan, J.~Quan, T.~Ramalho, A.~Grabska-Barwinska, et~al.
\newblock Overcoming catastrophic forgetting in neural networks.
\newblock \emph{Proceedings of the national academy of sciences}, 114\penalty0
  (13):\penalty0 3521--3526, 2017.

\bibitem[LeCun et~al.(1998)LeCun, Bottou, Bengio, and Haffner]{726791}
Y.~LeCun, L.~Bottou, Y.~Bengio, and P.~Haffner.
\newblock Gradient-based learning applied to document recognition.
\newblock \emph{Proceedings of the IEEE}, 86\penalty0 (11):\penalty0
  2278--2324, 1998.

\bibitem[L{\'e}cuyer et~al.(2019)L{\'e}cuyer, Spahn, Vodrahalli, Geambasu, and
  Hsu]{10.1145/3352020.3352032}
M.~L{\'e}cuyer, R.~Spahn, K.~Vodrahalli, R.~Geambasu, and D.~Hsu.
\newblock Privacy accounting and quality control in the sage differentially
  private ml platform.
\newblock In \emph{ACM Symposium on Operating Systems Principles}, pp.\
  181--195, 2019.

\bibitem[Liu et~al.(2019)Liu, Sun, and Fang]{Liu2019LifelongLF}
H.~Liu, F.~Sun, and B.~Fang.
\newblock Lifelong learning for heterogeneous multi-modal tasks.
\newblock In \emph{ICRA}, pp.\  6158--6164, 2019.

\bibitem[Lopez-Paz \& Ranzato(2017)Lopez-Paz and Ranzato]{NIPS2017_7225}
D.~Lopez-Paz and M.A. Ranzato.
\newblock Gradient episodic memory for continual learning.
\newblock \emph{NeurIPS}, 30, 2017.

\bibitem[Maschler et~al.(2021)Maschler, Pham, and
  Weyrich]{maschler2021regularization}
B.~Maschler, T.T.H. Pham, and M.~Weyrich.
\newblock Regularization-based continual learning for anomaly detection in
  discrete manufacturing.
\newblock \emph{Procedia CIRP}, 104:\penalty0 452--457, 2021.

\bibitem[Miotto et~al.(2016)Miotto, Li, Kidd, and Dudley]{citeulike:14040136}
R.~Miotto, L.~Li, B.A. Kidd, and J.T. Dudley.
\newblock Deep patient: an unsupervised representation to predict the future of
  patients from the electronic health records.
\newblock \emph{Scientific reports}, 6\penalty0 (1):\penalty0 1--10, 2016.

\bibitem[Ostapenko et~al.(2019)Ostapenko, Puscas, Klein, Jahnichen, and
  Nabi]{DBLP:conf/cvpr/OstapenkoPKJN19}
O.~Ostapenko, M.~Puscas, T.~Klein, P.~Jahnichen, and M.~Nabi.
\newblock Learning to remember: A synaptic plasticity driven framework for
  continual learning.
\newblock In \emph{CVPR}, pp.\  11321--11329, 2019.

\bibitem[Papernot et~al.(2016)Papernot, McDaniel, Sinha, and
  Wellman]{DBLP:journals/corr/PapernotMSW16}
N.~Papernot, P.~McDaniel, A.~Sinha, and M.~Wellman.
\newblock Towards the science of security and privacy in machine learning.
\newblock \emph{arXiv preprint arXiv:1611.03814}, 2016.

\bibitem[Papernot et~al.(2018)Papernot, Song, Mironov, Raghunathan, Talwar, and
  Erlingsson]{papernot2018scalable}
N.~Papernot, S.~Song, I.~Mironov, A.~Raghunathan, K.~Talwar, and
  {\'U}.~Erlingsson.
\newblock Scalable private learning with pate.
\newblock \emph{ICLR}, 2018.

\bibitem[Phan et~al.(2017{\natexlab{a}})Phan, Wu, and Dou]{PhanMLJ2017}
N.H. Phan, X.~Wu, and D.~Dou.
\newblock Preserving differential privacy in convolutional deep belief
  networks.
\newblock \emph{Machine learning}, 106\penalty0 (9):\penalty0 1681--1704,
  2017{\natexlab{a}}.

\bibitem[Phan et~al.(2017{\natexlab{b}})Phan, Wu, Hu, and Dou]{NHPhanICDM17}
N.H. Phan, X.~Wu, H.~Hu, and D.~Dou.
\newblock Adaptive laplace mechanism: Differential privacy preservation in deep
  learning.
\newblock In \emph{IEEE ICDM}, pp.\  385--394, 2017{\natexlab{b}}.

\bibitem[Phan et~al.(2019{\natexlab{a}})Phan, My, Devu, and
  Jin]{phan2019private}
N.H. Phan, T.~My, M.S. Devu, and R.~Jin.
\newblock Differentially private lifelong learning.
\newblock In \emph{Privacy in Machine Learning (PriML), NeurIPS'19 Workshop},
  2019{\natexlab{a}}.

\bibitem[Phan et~al.(2019{\natexlab{b}})Phan, Vu, Liu, Jin, Dou, Wu, and
  Thai]{PhanIJCAI}
N.H. Phan, M.~Vu, Y.~Liu, R.~Jin, D.~Dou, X.~Wu, and M.T. Thai.
\newblock Heterogeneous gaussian mechanism: Preserving differential privacy in
  deep learning with provable robustness.
\newblock \emph{IJCAI}, 2019{\natexlab{b}}.

\bibitem[Phan et~al.(2020)Phan, Thai, Hu, Jin, Sun, and
  Dou]{DBLP:journals/corr/abs-1903-09822}
N.H. Phan, M.T. Thai, H.~Hu, R.~Jin, T.~Sun, and D.~Dou.
\newblock Scalable differential privacy with certified robustness in
  adversarial learning.
\newblock In \emph{ICML}, pp.\  7683--7694, 2020.

\bibitem[Plis et~al.(2014)Plis, Hjelm, Salakhutdinov, Allen, Bockholt, Long,
  Johnson, Paulsen, Turner, and Calhoun]{10.3389/fnins.2014.00229}
S.M. Plis, D.R. Hjelm, R.~Salakhutdinov, E.A. Allen, H.J. Bockholt, J.D. Long,
  H.J. Johnson, J.S. Paulsen, J.A. Turner, and V.D. Calhoun.
\newblock Deep learning for neuroimaging: a validation study.
\newblock \emph{Frontiers in neuroscience}, 8:\penalty0 229, 2014.

\bibitem[Qu et~al.(2021)Qu, Rahmani, Xu, Williams, and Liu]{qu2021recent}
H.~Qu, H.~Rahmani, L.~Xu, B.~Williams, and J.~Liu.
\newblock Recent advances of continual learning in computer vision: An
  overview.
\newblock \emph{arXiv preprint arXiv:2109.11369}, 2021.

\bibitem[Rajasegaran et~al.(2020)Rajasegaran, Khan, Hayat, Khan, and
  Shah]{rajasegaran2020itaml}
J.~Rajasegaran, S.~Khan, M.~Hayat, F.S. Khan, and M.~Shah.
\newblock itaml: An incremental task-agnostic meta-learning approach.
\newblock In \emph{CVPR}, pp.\  13588--13597, 2020.

\bibitem[Rebuffi et~al.(2017)Rebuffi, Kolesnikov, Sperl, and Lampert]{8100070}
S.A. Rebuffi, A.~Kolesnikov, G.~Sperl, and C.H. Lampert.
\newblock icarl: Incremental classifier and representation learning.
\newblock In \emph{CVPR}, pp.\  2001--2010, 2017.

\bibitem[Riemer et~al.(2019)Riemer, Cases, Ajemian, Liu, Rish, Tu, and
  Tesauro]{riemer2018learning}
M.~Riemer, I.~Cases, R.~Ajemian, M.~Liu, I.~Rish, Y.~Tu, and G.~Tesauro.
\newblock Learning to learn without forgetting by maximizing transfer and
  minimizing interference.
\newblock \emph{ICLR}, 2019.

\bibitem[Roumia \& Steinhubl(2014)Roumia and Steinhubl]{Roumia2014}
M.~Roumia and S.~Steinhubl.
\newblock Improving cardiovascular outcomes using electronic health records.
\newblock \emph{Current cardiology reports}, 16\penalty0 (2):\penalty0 1--6,
  2014.

\bibitem[Shin et~al.(2017)Shin, Lee, Kim, and Kim]{10.5555/3294996.3295059}
H.~Shin, J.K. Lee, J.~Kim, and J.~Kim.
\newblock Continual learning with deep generative replay.
\newblock \emph{NeurIPS}, 30, 2017.

\bibitem[Shokri et~al.(2017)Shokri, Stronati, Song, and
  Shmatikov]{2016arXiv161005820S}
R.~Shokri, M.~Stronati, C.~Song, and V.~Shmatikov.
\newblock Membership inference attacks against machine learning models.
\newblock In \emph{IEEE Symposium on Security and Privacy (SP)}, pp.\  3--18,
  2017.

\bibitem[Tao et~al.(2020)Tao, Hong, Chang, Dong, Wei, and Gong]{tao2020fewshot}
X.~Tao, X.~Hong, X.~Chang, S.~Dong, X.~Wei, and Y.~Gong.
\newblock Few-shot class-incremental learning.
\newblock In \emph{CVPR}, pp.\  12183--12192, 2020.

\bibitem[Von~Oswald et~al.(2020)Von~Oswald, Henning, Sacramento, and
  Grewe]{von2019continual}
J.~Von~Oswald, C.~Henning, J.~Sacramento, and B.F. Grewe.
\newblock Continual learning with hypernetworks.
\newblock \emph{ICLR}, 2020.

\bibitem[Wang et~al.(2015)Wang, Si, and Wu]{DBLP:conf/ijcai/SW15}
Y.~Wang, C.~Si, and X.~Wu.
\newblock Regression model fitting under differential privacy and model
  inversion attack.
\newblock In \emph{IJCAI}, 2015.

\bibitem[Wu et~al.(2018)Wu, Herranz, Liu, van~de Weijer, Raducanu,
  et~al.]{NIPS2018_7836}
C.~Wu, L.~Herranz, X.~Liu, J.~van~de Weijer, B.~Raducanu, et~al.
\newblock Memory replay gans: Learning to generate new categories without
  forgetting.
\newblock \emph{NeurIPS}, 31, 2018.

\bibitem[Wu et~al.(2010)Wu, Roy, and Stewart]{citeulike:7685411}
J.~Wu, J.~Roy, and W.F. Stewart.
\newblock Prediction modeling using ehr data: challenges, strategies, and a
  comparison of machine learning approaches.
\newblock \emph{Medical care}, pp.\  S106--S113, 2010.

\bibitem[Ye \& Bors(2020)Ye and Bors]{ye2020learning}
F.~Ye and A.G. Bors.
\newblock Learning latent representations across multiple data domains using
  lifelong vaegan.
\newblock In \emph{European Conference on Computer Vision}, pp.\  777--795,
  2020.

\bibitem[Yoon et~al.(2020)Yoon, Kim, Seo, and Moon]{yoon2020xtarnet}
S.W. Yoon, D.Y. Kim, J.~Seo, and J.~Moon.
\newblock Xtarnet: Learning to extract task-adaptive representation for
  incremental few-shot learning.
\newblock In \emph{International Conference on Machine Learning}, pp.\
  10852--10860, 2020.

\bibitem[Zhang et~al.(2012)Zhang, Zhang, Xiao, Yang, and
  Winslett]{zhang2012functional}
J.~Zhang, Z.~Zhang, X.~Xiao, Y.~Yang, and M.~Winslett.
\newblock Functional mechanism: regression analysis under differential privacy.
\newblock \emph{PVLDB}, pp.\  21364–--1375, 2012.

\end{thebibliography}
\bibliographystyle{collas2022_conference}

\appendix

\onecolumn

\section{Proof of Theorem \ref{PRTheorem}}

\begin{proof} Let us denote $A_{1:i}$ as $A_1, \ldots, A_i$, we have: 
\begin{align}
\scriptsize
\nonumber c(\theta^\tau; A_\tau, \{\theta^i\}_{i<\tau}, \mathsf{data}_\tau, \mathsf{data}'_\tau) &= \log \frac{Pr[A_\tau(\{\theta^i\}_{i<\tau}, \mathsf{data}_\tau) = \theta^\tau]}{Pr[A_\tau(\{\theta^i\}_{i<\tau}, \mathsf{data}'_\tau) = \theta^\tau]} \nonumber \\
&= \log \prod_{i = 1}^{\tau} \frac{Pr[A_{i}(\theta^{i-1}, \mathsf{data}_i) = \theta^i \lvert A_{1:i-1}(\{\theta^{j}\}_{j < i-1},\mathsf{data}_{1:i-1}) = \theta^{1:i-1}]}{Pr[A_{i}(\theta^{i-1}, \mathsf{data}'_i) = \theta^i \lvert A_{1:i-1}(\{\theta^{j}\}_{j < i-1},\mathsf{data}'_{1:i-1}) = \theta^{1:i-1}]} \nonumber \\
&= \sum_{i = 1}^{\tau} \log \frac{Pr[A_{i}(\theta^{i-1}, \mathsf{data}_i) = \theta^i \lvert A_{1:i-1}(\{\theta^{j}\}_{j < i-1},\mathsf{data}_{1:i-1}) = \theta^{1:i-1}]}{Pr[A_{i}(\theta^{i-1}, \mathsf{data}'_i) = \theta^i \lvert A_{1:i-1}(\{\theta^{j}\}_{j < i-1},\mathsf{data}'_{1:i-1}) = \theta^{1:i-1}]} \nonumber \\
&= \sum_{i = 1}^{\tau} c(\theta^i; A_i, \{\theta^j\}_{j<i}, \mathsf{data}_i, \mathsf{data}'_i) \nonumber
\end{align}
Consequently, Theorem \ref{PRTheorem} does hold.
\end{proof}

\section{Proof of Theorem \ref{OverallL2DP}} \label{ProofT2}

\begin{proof}
$\forall \tau \in \mathbf{T}$, let $\overline{D}_\tau$ and $\overline{D}'_\tau$ be neighboring datasets differing at most one tuple $x_e \in \overline{D}_\tau$ and $x'_e \in \overline{D}'_\tau$, and any two neighboring episodic memories $\mathbb{M}_\tau$ and $\mathbb{M}'_\tau$. Let us denote Alg. \ref{DPL2M-Dataset} as the mechanism $A$ in Definition \ref{LifelongDP}. We first show that Alg. \ref{DPL2M-Dataset} achieves typical DP protection. $\forall \tau$ and $D_{ref}$, we have that
\begin{align}
& Pr\big[A(\{\theta^i\}_{i < \tau}, \mathsf{data}_\tau) = \theta^\tau \big] \\&=  Pr\big(\overline{\mathcal{R}}_{\overline{D}_\tau}(\theta^{\tau-1}_1)\big) Pr\big(\overline{D}_\tau \big) Pr\big(\overline{\mathcal{L}}_{\overline{D}_\tau}(\theta^{\tau-1}_2)\big)  \times Pr\big(\overline{\mathcal{R}}_{\overline{D}_{ref}}(\theta^{\tau-1}_1)\big) Pr\big(\overline{D}_{ref} \big) Pr\big(\overline{\mathcal{L}}_{\overline{D}_{ref}}(\theta^{\tau-1}_2)\big) \nonumber
\end{align}

Therefore, we further have
\begin{align}
\nonumber & \frac{Pr\big[A(\{\theta^i\}_{i < \tau}, \mathsf{data}_\tau) = \theta^\tau \big]}{Pr\big[A(\{\theta^i\}_{i < \tau}, \mathsf{data}'_\tau) = \theta^\tau \big]} \\&=
\frac{Pr\big(\overline{\mathcal{R}}_{\overline{D}_\tau}(\theta^{\tau-1}_1)\big)}{Pr\big(\overline{\mathcal{R}}_{\overline{D}'_\tau}(\theta^{\tau-1}_1)\big)} \frac{Pr\big(\overline{D}_\tau \big)}{Pr\big(\overline{D}'_\tau \big)} \frac{Pr\big(\overline{\mathcal{L}}_{\overline{D}_\tau}(\theta^{\tau-1}_2)\big)}{Pr\big(\overline{\mathcal{L}}_{\overline{D}'_\tau}(\theta^{\tau-1}_2)\big)} \times \frac{Pr\big(\overline{\mathcal{R}}_{\overline{D}_{ref}}(\theta^{\tau-1}_1)\big)}{Pr\big(\overline{\mathcal{R}}_{\overline{D}'_{ref}}(\theta^{\tau-1}_1)\big)} \frac{Pr\big(\overline{D}_{ref} \big)}{Pr\big(\overline{D}'_{ref} \big)} \frac{Pr\big(\overline{\mathcal{L}}_{\overline{D}_{ref}}(\theta^{\tau-1}_2)\big)} {Pr\big(\overline{\mathcal{L}}_{\overline{D}'_{ref}}(\theta^{\tau-1}_2)\big)}
\label{PCond1}
\end{align}

In addition, we also have that:
\begin{equation}
\exists! \overline{D}_\tau \in \overline{\mathcal{D}} \textit{ s.t. } x_e \in \overline{D}_\tau \textit{ and } \exists! \overline{D}'_\tau \in \overline{\mathcal{D}}' \textit{ s.t. } x'_e \in \overline{D}'_\tau
\label{PCond2}
\end{equation}
where $\overline{\mathcal{D}} = \{\overline{D}_1, \ldots, \overline{D}_m\}$.

Together with Eq. \ref{PCond2}, by having disjoint and fixed datasets in the episodic memory, we have that: 
\begin{equation}
(x_e \in \overline{D}_\tau \textit{\ or \ } x_e \in \overline{D}_{ref}), \textit{\ but \ } (x_e \in \overline{D}_\tau \textit{\ and \ } x_e \in \overline{D}_{ref})
\label{PCond3}
\end{equation}

Without loss of the generality, we can assume that $x_e \in \overline{D}_\tau$: Eqs. \ref{PCond1} - \ref{PCond3} $\Rightarrow$ 
\begin{align}
\frac{Pr\big[A(\{\theta^i\}_{i < \tau}, \mathsf{data}_\tau) = \theta^\tau \big]}{Pr\big[A(\{\theta^i\}_{i < \tau}, \mathsf{data}'_\tau) = \theta^\tau \big]}  = \frac{Pr\big(\overline{\mathcal{R}}_{\overline{D}_\tau}(\theta^{\tau-1}_1)\big)}{Pr\big(\overline{\mathcal{R}}_{\overline{D}'_\tau}(\theta^{\tau-1}_1)\big)} \frac{Pr\big(\overline{D}_\tau \big)}{Pr\big(\overline{D}'_\tau \big)} \frac{Pr\big(\overline{\mathcal{L}}_{\overline{D}_\tau}(\theta^{\tau-1}_2)\big)}{Pr\big(\overline{\mathcal{L}}_{\overline{D}'_\tau}(\theta^{\tau-1}_2)\big)} 
\leq (\epsilon_1 + \epsilon_1/\gamma_{\mathbf{x}} + \epsilon_1/\gamma+\epsilon_2)  \label{Consistency1}
\end{align}

This is also true when $x_e \in \overline{D}_{ref}$ \textit{and} $x_e \not\in \overline{D}_\tau$.

As a result, we have
\begin{equation}
\forall \tau \in [1, m]: \frac{Pr\big[A(\{\theta^i\}_{i < \tau}, \mathsf{data}_\tau) = \theta^\tau \big]}{Pr\big[A(\{\theta^i\}_{i < \tau}, \mathsf{data}'_\tau) = \theta^\tau \big]} 
\leq (\epsilon_1 + \epsilon_1/\gamma_{\mathbf{x}} + \epsilon_1/\gamma+\epsilon_2)
\label{LocalDP-Result}
\end{equation}

After one training step, $\overline{D}_\tau$ will be placed into the episodic memory $\mathbb{M}_\tau$ to create the memory $\mathbb{M}_{\tau + 1}$. In the next training task, $\overline{D}_\tau$ can be randomly selected to compute the episodic gradient $g_{ref}$. This computation does not incur any additional privacy budget consumption for the dataset $\overline{D}_\tau$, by applying the Theorem 4 in \citep{DBLP:journals/corr/abs-1903-09822}, which allows us to \textit{compute gradients across an unlimited number of training steps} using $\overline{\mathcal{R}}_{\overline{D}_\tau}(\theta^{\tau-1}_1)$ and $\overline{\mathcal{L}}_{\overline{D}_\tau}(\theta^{\tau-1}_2)$. Therefore, if the same privacy budget is used for all the training tasks in $\mathbf{T}$, we will have only one privacy loss for every tuple in all the tasks. The optimization in one task does not affect the DP guarantee of any other tasks. Consequently, we have
\begin{align}
\nexists \epsilon' < (\epsilon_1 + \epsilon_1/\gamma_{\mathbf{x}} + \epsilon_1/\gamma+\epsilon_2), \exists i \leq m  \text{ s.t. } Pr\big[A(\{\theta^j\}_{j < i}, \mathsf{data}_i) = \theta^i \big]  \leq e^{\epsilon'} Pr\big[A(\{\theta^j\}_{j < i}, \mathsf{data}'_i) = \theta^i \big] \label{FairDP-Result}
\end{align}

Eq. \ref{FairDP-Result} can be further used to prove the Lifelong DP protection. Given $\mathsf{data}_m$ where $M_t = \overline{D}_t$ in Alg. \ref{DPL2M-Dataset}, we have that
\begin{equation}
Pr\big[A(\mathsf{data}_m) = \{\theta^i\}_{i \in [1, m]} \big] 
= \prod_{i = 1}^m Pr\big[A(\{\theta^j\}_{j < i}, \mathsf{data}_i) = \theta^i \big]
\end{equation}

Therefore, we have 
\begin{align}
& \frac{Pr\big[A(\mathsf{data}_m) = \{\theta^i\}_{i \in [1, m]} \big]}{Pr\big[A(\mathsf{data}'_m) = \{\theta^i\}_{i \in [1, m]} \big]} = \prod_{i = 1}^m \frac{Pr\big[A(\{\theta^j\}_{j < i}, \mathsf{data}_i) = \theta^i \big]}{Pr\big[A(\{\theta^j\}_{j < i}, \mathsf{data}'_i) = \theta^i \big]} \nonumber \\
& = \prod_{i = 1}^m \Big[\frac{Pr\big(\overline{\mathcal{R}}_{\overline{D}_i}(\theta^{i-1}_1)\big)}{Pr\big(\overline{\mathcal{R}}_{\overline{D}'_i}(\theta^{i-1}_1)\big)} \frac{Pr\big(\overline{D}_i \big)}{Pr\big(\overline{D}'_i \big)} \frac{Pr\big(\overline{\mathcal{L}}_{\overline{D}_i}(\theta^{i-1}_2)\big)}{Pr\big(\overline{\mathcal{L}}_{\overline{D}'_i}(\theta^{i-1}_2)\big)} \times  \frac{Pr\big(\overline{\mathcal{R}}_{\overline{D}_{ref}^i}(\theta^{i-1}_1)\big)}{Pr\big(\overline{\mathcal{R}}_{\overline{D}_{ref}^{i'}}(\theta^{i-1}_1)\big)} \frac{Pr\big(\overline{D}_{ref}^i \big)}{Pr\big(\overline{D}_{ref}^{i'} \big)} \frac{Pr\big(\overline{\mathcal{L}}_{\overline{D}_{ref}^i}(\theta^{i-1}_2)\big)} {Pr\big(\overline{\mathcal{L}}_{\overline{D}_{ref}^{i'}}(\theta^{i-1}_2)\big)} \Big]  \label{globalCond1} 
\end{align}
where $\mathsf{data}'_m = \{\overline{\mathcal{D}}, \{M'_i\}_{i \in [1, m]}\}$, and $M'_i = \overline{D}'_i$ in Alg. \ref{DPL2M-Dataset}.

Since all the datasets are non-overlapping, i.e., $\cap_{i \in [1, m]}D_i = \emptyset$, given an arbitrary tuple $x_e$, we have that 
\begin{equation}
\exists! \overline{D}_\tau \in \overline{\mathcal{D}} \textit{ s.t. } x_e \in \overline{D}_\tau \textit{ and } \exists! \overline{D}'_\tau \in \overline{\mathcal{D}}' \textit{ s.t. } x'_e \in \overline{D}'_\tau
\label{globalCond2}
\end{equation}

Thus, the optimization of $\{\theta_{1}^{i}, \theta_{2}^{i}\} = \arg\min_{\theta_1, \theta_2} [\overline{\mathcal{R}}_{\overline{D}_i}(\theta^{i-1}_1) + \overline{\mathcal{L}}_{\overline{D}_i}(\theta^{i-1}_2)]$ for any other task $i$ different from $\tau$ does not affect the privacy protection of $x_e$ in $\overline{\mathcal{D}}$. From Eqs. \ref{globalCond1} and \ref{globalCond2}, we have
\begin{align}
&\nonumber \frac{Pr\big[A(\mathsf{data}_m) = \{\theta^i\}_{i \in [1, m]} \big]}{Pr\big[A(\mathsf{data}'_m) = \{\theta^i\}_{i \in [1, m]} \big]} \\
&=
\frac{Pr\big(\overline{\mathcal{R}}_{\overline{D}_\tau}(\theta^{\tau-1}_1)\big)}{Pr\big(\overline{\mathcal{R}}_{\overline{D}'_\tau}(\theta^{\tau-1}_1)\big)} \frac{Pr\big(\overline{D}_\tau \big)}{Pr\big(\overline{D}'_\tau \big)} \frac{Pr\big(\overline{\mathcal{L}}_{\overline{D}_\tau}(\theta^{\tau-1}_2)\big)}{Pr\big(\overline{\mathcal{L}}_{\overline{D}'_\tau}(\theta^{\tau-1}_2)\big)} \times \prod_{i = 1}^m \frac{Pr\big(\overline{\mathcal{R}}_{\overline{D}_{ref}^i}(\theta^{i-1}_1)\big)}{Pr\big(\overline{\mathcal{R}}_{\overline{D}_{ref}^{i'}}(\theta^{i-1}_1)\big)} \frac{Pr\big(\overline{D}_{ref}^i \big)}{Pr\big(\overline{D}_{ref}^{i'} \big)} \frac{Pr\big(\overline{\mathcal{L}}_{\overline{D}_{ref}^i}(\theta^{i-1}_2)\big)} {Pr\big(\overline{\mathcal{L}}_{\overline{D}_{ref}^{i'}}(\theta^{i-1}_2)\big)}
\label{globalCond3}
\end{align}

The worse privacy leakage case to $x_e$ is that $\overline{D}_\tau$ is used in every $\overline{D}^i_{ref}$, i.e., $\tau = 1$ and $\forall i \in [2, m]: \overline{D}^i_{ref} = \overline{D}_\tau$, with $\overline{D}^1_{ref} = \emptyset$. Meanwhile, the least privacy leakage case to $x_e$ is that $\overline{D}_\tau$ is not used in any $\overline{D}^i_{ref}$, i.e., $\forall i \in [2, m]: \overline{D}^i_{ref} \neq \overline{D}_\tau$, with $\overline{D}^1_{ref} = \emptyset$. In order to bound the privacy loss, we consider the worse case; therefore, from Eq. \ref{globalCond3}, we further have that
\begin{equation}
\frac{Pr\big[A(\mathsf{data}_m) = \{\theta^i\}_{i \in [1, m]} \big]}{Pr\big[A(\mathsf{data}'_m) = \{\theta^i\}_{i \in [1, m]} \big]} 
\leq \prod_{i = 1}^m \frac{Pr\big(\overline{\mathcal{R}}_{\overline{D}_\tau}(\theta^{i - 1}_1)\big)}{Pr\big(\overline{\mathcal{R}}_{\overline{D}'_\tau}(\theta^{i - 1}_1)\big)} \frac{Pr\big(\overline{D}_\tau \big)}{Pr\big(\overline{D}'_\tau \big)} \frac{Pr\big(\overline{\mathcal{L}}_{\overline{D}_\tau}(\theta^{i - 1}_2)\big)}{Pr\big(\overline{\mathcal{L}}_{\overline{D}'_\tau}(\theta^{i - 1}_2)\big)}
\label{globalCond4}
\end{equation}

Eq. \ref{globalCond4} is equivalent to the continuously training of our model by optimizing $\overline{\mathcal{R}}$ and $\overline{\mathcal{L}}$ with $\overline{D}_\tau$ used as both the current task and the episodic memory, across $m$ steps. By following the Theorem 4 in \citep{DBLP:journals/corr/abs-1903-09822}, the privacy budget is not accumulated across training steps. Therefore, we have that 
\begin{align}
\forall m \in [1, \infty): & \frac{Pr\big[A(\mathsf{data}_m) = \{\theta^i\}_{i \in [1, m]} \big]}{Pr\big[A(\mathsf{data}'_m) = \{\theta^i\}_{i \in [1, m]} \big]} \nonumber  \leq \prod_{i = 1}^m \frac{Pr\big(\overline{\mathcal{R}}_{\overline{D}_\tau}(\theta^{i - 1}_1)\big)}{Pr\big(\overline{\mathcal{R}}_{\overline{D}'_\tau}(\theta^{i - 1}_1)\big)} \frac{Pr\big(\overline{D}_\tau \big)}{Pr\big(\overline{D}'_\tau \big)} \frac{Pr\big(\overline{\mathcal{L}}_{\overline{D}_\tau}(\theta^{i - 1}_2)\big)}{Pr\big(\overline{\mathcal{L}}_{\overline{D}'_\tau}(\theta^{i - 1}_2)\big)} \nonumber 
\\
& = \frac{Pr\big(\overline{\mathcal{R}}_{\overline{D}_\tau}(\theta_1)\big)}{Pr\big(\overline{\mathcal{R}}_{\overline{D}'_\tau}(\theta_1)\big)} \frac{Pr\big(\overline{D}_\tau \big)}{Pr\big(\overline{D}'_\tau \big)} \frac{Pr\big(\overline{\mathcal{L}}_{\overline{D}_\tau}(\theta_2)\big)}{Pr\big(\overline{\mathcal{L}}_{\overline{D}'_\tau}(\theta_2)\big)}   \leq (\epsilon_1 + \epsilon_1/\gamma_{\mathbf{x}} + \epsilon_1/\gamma+\epsilon_2)
\label{GlobalDP-Result}
\end{align}

In the least privacy leakage case, we have that
\begin{equation}
\forall \tau \leq m: 
\frac{Pr\big[A(\mathsf{data}_\tau) = \{\theta^i\}_{i \in [1, \tau]} \big]}{Pr\big[A(\mathsf{data}'_\tau) = \{\theta^i\}_{i \in [1, \tau]} \big]} 
 \geq \frac{Pr\big[A(\{\theta^i\}_{i < \tau}, \mathsf{data}_\tau) = \theta^\tau \big]}{Pr\big[A(\{\theta^i\}_{i < \tau}, \mathsf{data}'_\tau) = \theta^\tau \big]} \geq (\epsilon_1 + \epsilon_1/\gamma_{\mathbf{x}} + \epsilon_1/\gamma+\epsilon_2)
\end{equation}

As a result, we have that
\begin{equation}
\nexists (\epsilon' < \epsilon, \tau \leq m): Pr\big[A(\mathsf{data}_\tau) = \{\theta^i\}_{i \in [1, \tau]} \big] \leq e^{\epsilon'} Pr\big[A(\mathsf{data}'_\tau) = \{\theta^i\}_{i \in [1, \tau]} \big]
\label{Consistency-Result}
\end{equation}
where $\epsilon = (\epsilon_1 + \epsilon_1/\gamma_{\mathbf{x}} + \epsilon_1/\gamma+\epsilon_2)$.

From Eqs. \ref{GlobalDP-Result} and \ref{Consistency-Result}, we have that Alg. \ref{DPL2M-Dataset} achieves $(\epsilon_1 + \epsilon_1/\gamma_{\mathbf{x}} + \epsilon_1/\gamma+\epsilon_2)$-Lifelong DP in learning $\{\theta^i\}_{i \in [1, m]} = \{\theta^{i}_1, \theta^{i}_2\}_{i \in [1, m]}$.
Consequently, Theorem \ref{OverallL2DP} does hold.
\end{proof}

\section{L2DP-ML with Streaming Batch Training}
\label{L2DP-ML with Streaming}

\begin{algorithm}[h]
\caption{\textbf{L2DP-ML with Streaming Batch Training}}
\label{DPL2M} 
\KwIn{$\mathbf{T}$=$\{t_i\}_{i \in [1,m]}$, $\{D_i\}_{i \in [1,m]}$, batch size $\lambda$, privacy budgets: $\epsilon_1$ and $\epsilon_2$, learning rate $\varrho$}
\KwOut{$(\epsilon_1 + \epsilon_1/\gamma_{\mathbf{x}} + \epsilon_1/\gamma+\epsilon_2)$-Lifelong DP parameters $\{\theta^i\}_{i \in [1, m]} = \{\theta^{i}_1, \theta^{i}_2\}_{i \in [1, m]}$}
\begin{algorithmic}[1]
\STATE \textbf{Draw Noise} $\chi_1 \leftarrow [Lap(\frac{\Delta_{\widetilde{\mathcal{R}}}}{\epsilon_1})]^{d}$, $\chi_2 \leftarrow [Lap(\frac{\Delta_{\widetilde{\mathcal{R}}}}{\epsilon_1})]^{\beta}$, $\chi_3 \leftarrow [Lap(\frac{\Delta_{\widetilde{\mathcal{L}}}}{\epsilon_2})]^{\lvert \mathbf{h}_\pi\lvert }$
\STATE \textbf{Randomly Initialize} $\theta = \{\theta_1, \theta_2\}$, $\mathbb{M}_1 = \emptyset$, $\forall \tau \in \mathbf{T}: \overline{D}_\tau = \{\overline{x}_r \leftarrow x_{r} + \frac{\chi_1}{\lambda}\}_{x_r \in D_\tau}$, hidden layers $\{\mathbf{h}_1 + \frac{2\chi_2}{\lambda},\dots, \mathbf{h}_\pi\}$, where $\mathbf{h}_\pi$ is the last hidden layer
\FOR{$\tau \in \mathbf{T}$}
    \STATE $\mathbf{B} = \{B_1, \ldots, B_{n}\}$ s.t. $\forall B \in \mathbf{B}: B$ is a random batch with the size $s$, $B_1 \cap \ldots \cap B_{n} = \emptyset$, and $B_1 \cup \ldots \cup B_{n} = \overline{D}_\tau$
    \FOR{$B \in \mathbf{B}$}
        \IF{$\tau == 0$}
        	  \STATE \textbf{Compute Gradients: }
        	  \STATE $g \leftarrow \{\nabla_{\theta_1}\overline{\mathcal{R}}_{B}(\theta^{\tau-1}_1), \nabla_{\theta_2} \overline{\mathcal{L}}_{B}(\theta^{\tau-1}_2)\}$ with the noise $\frac{\chi_3}{\lambda}$
	        \STATE \textbf{Descent: } $\{\theta^{\tau}_1, \theta^{\tau}_2\} \leftarrow \{\theta^{\tau-1}_1, \theta^{\tau-1}_2\} - \varrho g$
	    \ELSE
	        \STATE \textbf{Select} a batch $B_e$ randomly from a set of batches in episodic memory $\mathbb{M}_\tau$
	        \STATE \textbf{Compute Gradients: }
	        \STATE $g \leftarrow \{\nabla_{\theta_1}\overline{\mathcal{R}}_{B}(\theta^{\tau-1}_1), \nabla_{\theta_2} \overline{\mathcal{L}}_{B}(\theta^{\tau-1}_2)\}$ with the noise $\frac{\chi_3}{\lambda}$
	        \STATE $g_{ref} \leftarrow \{\nabla_{\theta_1}\overline{\mathcal{R}}_{B_e}(\theta^{\tau-1}_1), \nabla_{\theta_2} \overline{\mathcal{L}}_{B_e}(\theta^{\tau-1}_2)\}$ with the noise $\frac{\chi_3}{\lambda}$
	        \STATE $\tilde{g} \leftarrow g - \frac{g^{\top}g_{ref}}{g^{\top}_{ref}g_{ref}}g_{ref}$
	        \STATE \textbf{Descent: } $\{\theta^{\tau}_1, \theta^{\tau}_2\} \leftarrow \{\theta^{\tau-1}_1, \theta^{\tau-1}_2\} - \varrho \tilde{g}$
	    \ENDIF
    \ENDFOR
    \STATE \textbf{Randomly Select} a batch $B \in \mathbf{B}$
    \STATE $\mathbb{M}_{\tau} \leftarrow \mathbb{M}_{\tau-1} \cup B$
\ENDFOR
\end{algorithmic} 
\end{algorithm}

\section{HARW Dataset} \label{HARWData}

\textbf{Data Collection.} We utilize Android smartphones to collect smartphone sensor data ``in the wild" from university students as subjects for the following reasons: \textbf{(1)} University students should have relatively good access to the smartphones and related technologies; \textbf{(2)} University students should be more credible and easier to be motivated than other sources (e.g., recruiting test subjects on crowd-sourcing websites); and \textbf{(3)} It will be easier for our team to recruit and distribute rewards to students. 
We launched two data collection runs at two universities for three months each. During the course of three months, we let the participants to collect data and labels by themselves (in the wild), and only intervene through reminding emails if we saw a decline in the amount of daily activities. A total of 116 participants were recorded after the two data collection runs. 

\textbf{Data Processing.} For the demonstration purpose of this paper, we use only accelerometer data. Our data processing consists of the following steps:
\textbf{(1)} Any duplicated data points (e.g., data points that have the same timestamp) are merged by taking the average of their sensor values;
\textbf{(2)} Using 300 milliseconds as the threshold, continuous data sessions are identified and separated by breaking up the data sequences at any gap that is larger than the threshold;
\textbf{(3)} Data sessions that have unstable or unsuitable sampling rates are filtered out. We only keep the data sessions that have a stable sampling rate of 5Hz, 10Hz, 20Hz, or 50Hz; 
\textbf{(4)} The label sessions that are associated with each data session (if any) are identified from the raw labels. Note that the label sessions are also filtered with the following two criteria to ensure good quality: (a) The first 10 seconds and the last 10 seconds of each label session are trimmed, due to the fact that users were likely operating the phone during these time periods; (b) Any label session longer than 30 minutes is trimmed down to 30 minutes, in order to mitigate the potential inaccurate labels due to users’ negligence (forgot to turn off labeling); and 
\textbf{(5)} We sample data segments at the size of 100 data points with sliding windows. Different overlapping percentages were used for different classes and different sampling rates. The majority classes have 25\% overlapping to reduce the number of data segments, while the minority classes have up to 90\% overlapping to increase the available data segments. The same principle is applied to sessions with different sampling rates. 
We sample 15\% of data for testing, while the rest are used for training (Table \ref{tab:dataset}).

\textbf{Data Normalization.} In our L2DP-ML models, we normalize the accelerometer data with the following steps:
\textbf{(1)} We compute the mean and variance of each axis (i.e., $X$, $Y$, and $Z$) using only training data to avoid information leakage from the training phase to the testing phase. Then, both training and testing data are normalized with z-score, based on the mean and variance computed from training data;
\textbf{(2)} Based on this, we clip the values in between $[min, max] = [-2, 2]$ for each axis, which covers at least 90\% of possible data values; and
\textbf{(3)} Finally, all values are linearly scaled to $[-1, 1]$ to finish the normalization process, as
$
x = 2 \times [\frac{x - min}{max - min} - 1/2]
$. 

\begin{table}[t]
\caption{Statistics of the HARW dataset.} 
\label{tab:dataset}
\centering
\begin{tabular}{|l|l|l|l|}
\hline
Class            & Description                 & N training & N testing \\ \hline
Walking                     & Walking                     & 49376                            & 8599                            \\ \hline
Sitting                     & Exclude in vehicle & 52448                            & 8744                            \\ \hline
In-Vehicle, Car             & Driving, sitting          & 49536                            & 8586                            \\ \hline
Cycling                    &                             & 14336                            & 2537                            \\ \hline
Workout, Running            &                             & 1984                             & 319                             \\ \hline
\multicolumn{4}{|l|}{*All classes exclude phone position   = ``Table"}                                                             \\ \hline
\end{tabular} 
\end{table}

In the HARW dataset, each data tuple includes 100 values $\times$ 3 channels of the accelerometer sensor, i.e., 300 values in total as a model input. The classification output includes five classes of human activities, i.e., walking, sitting, in car, cycling, and running (Table \ref{tab:dataset}, Appx. \ref{HARWData}). Given 20Hz, 5Hz, 10Hz, and 50Hz tasks, we correspondingly have 881, 7553, 621, and 156,033 data points in training and 159, 1,297, 124, and 27,134 data points in testing. 

\textbf{Baseline Model Performance.} We conducted experiments on the HARW dataset in a centralized training on the whole dataset including all the data sampling rates using following baselines: 1) CNN-based model with the numbers of convolution-channels set to 32, 64, 128, denoted as CNN-32, CNN-64, CNN-128, respectively; 2) Bidirectional LSTM (BiLSTM); and 3) CNN-based models proposed by  \citet{ignatov2018real}, with additional features (CNN-Ig) and without additional features (CNN-Ig-featureless) using the Ignatov's recommended settings in \citet{ignatov2018real}.


\begin{table}[t]
\caption{Baseline results on the HARW dataset.} 
\label{tab:baseline}
\centering
\begin{tabular}{|l|l|l|l|}
\hline
Model         & Accuracy ($\%$)  \\ \hline
CNN-32                   & 81.86                        \\ \hline
CNN-64                  & 82.49                       \\ \hline
CNN-128                    & 82.62                     \\ \hline
BiLSTM                   &  78.68                    \\ \hline
CNN-Ig                    & 76.39                   \\ \hline
CNN-Ig-featureless                  & 77.08           
\\ \hline
\end{tabular} 
\end{table}

As in Table \ref{tab:baseline}, our model trained on each task independently achieves competitive results with these baselines under a rigorous DP budget ($\epsilon = 0.2$), i.e., 77\%, 76\%, 75\%, 58\%, on the 5Hz, 10Hz, 20Hz, and 50Hz learning tasks respectively. Although the number of 50Hz training data points is larger than other tasks, the data labels are noisy and collected in short-time periods due to the limited computational resources on mobile devices; thus, the model performance in the 50Hz learning task is lower.

\section{Hyper-parameter Grid-Search and Supplemental Results} \label{Hyper-parameter Search}

\textbf{Model Configuration.} In the permuted MNIST and the Split MNIST datasets, we used three convolutional layers (32, 64, and 96 features). Each hidden neuron connects with a 5x5 unit patch. A fully-connected layer has 512 units. In the permuted CIFAR-10 and the Split CIFAR-10/100 datasets, we used a Resnet-18 network (64, 64, 128, 128, and 160 features) with kernels (4, 3, 3, 3, and 3). One fully-connected layer has 256 neurons.
In the HARW dataset, we used three convolutional layers (32, 64, and 96 features). Each hidden neuron connects with a 2x2 unit patch. A fully-connected layer has 128 units.

In the Split CIFAR-10 and CIFAR-100 setting, there are $11$ tasks, in which 
the first task is the full CIFAR-10 classification task, and the remaining $10$ tasks consist of splits from the CIFAR-100 dataset. Each split contains $10$ classes from the CIFAR-100. We adopt this approach from \citep{von2019continual}. In the Split MNIST setting, there are $5$ tasks, in which each task consists of $2$ classes from the MNIST dataset. There is no overlapping classes between tasks in the Split CIFAR-10 and CIFAR-100, and in the Split MNIST. 

In order to be fair in comparison with the L2DP-ML and A-gem mechanisms, we conducted experiments over a wide range of privacy hyper-parameters such as privacy budget ($\epsilon$), noise scale ($z$), and sensitivity to select the best hyper-parameters in NaiveGaussian mechanism in our experiments. The search ranges and their results (i.e., average accuracy over all tasks) are provided in Table~\ref{tab:cifar-10}. 
We reported the best results, i.e., highest average accuracy over all tasks, of the hyper-parameter grid-search experiments.


\begin{figure*}[h]
  \centering
\subfloat[]{\label{PvalueMNISTCIFARa}\includegraphics[scale=0.15]{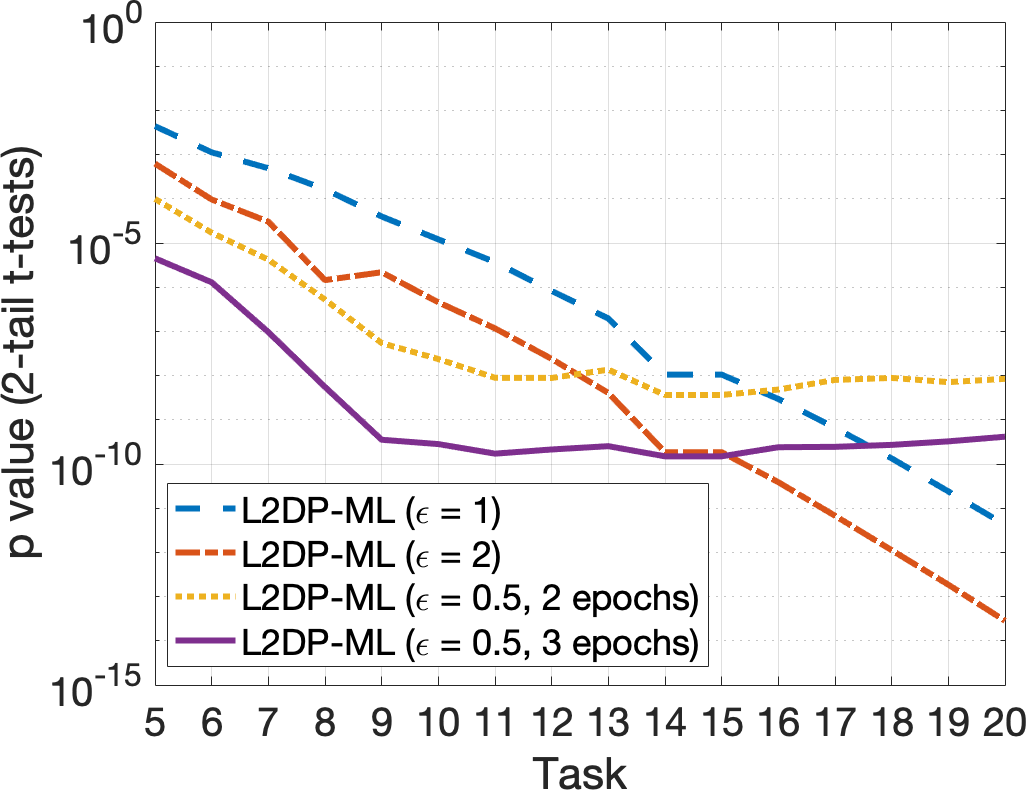}}\hspace*{0.1cm}
\subfloat[]{\label{PvalueMNISTCIFARb}\includegraphics[scale=0.15]{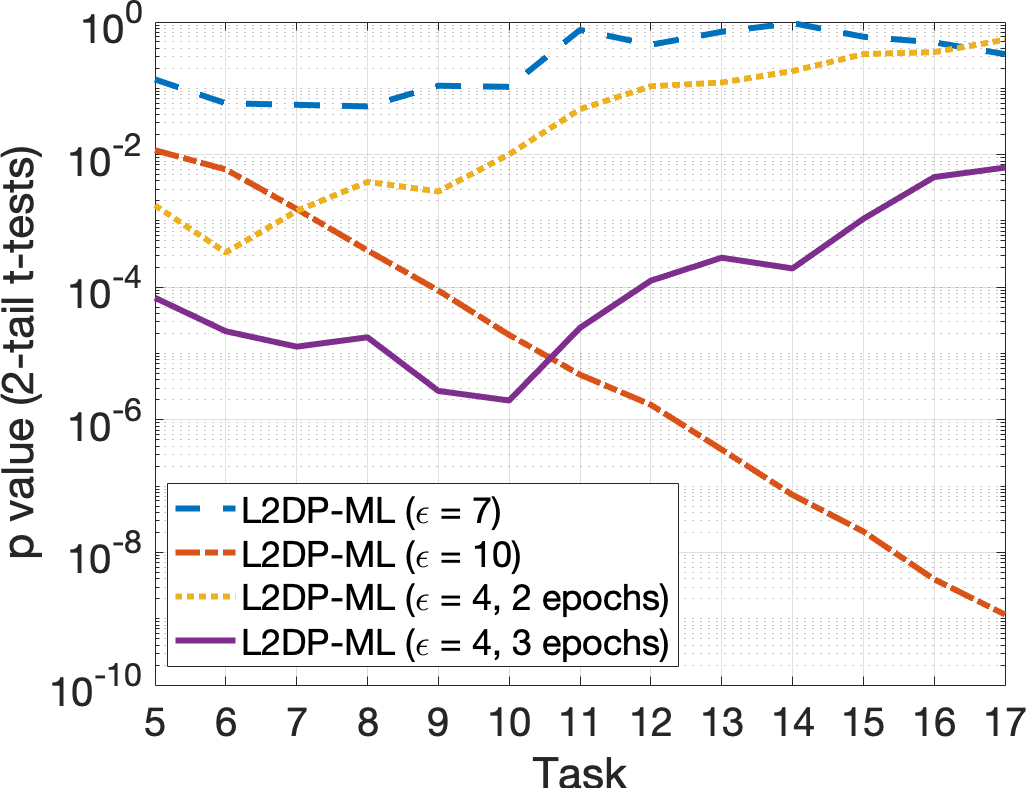}}
\hspace*{0.1cm}
\subfloat[]{\label{PvalueMNISTCIFARc}\includegraphics[scale=0.15]{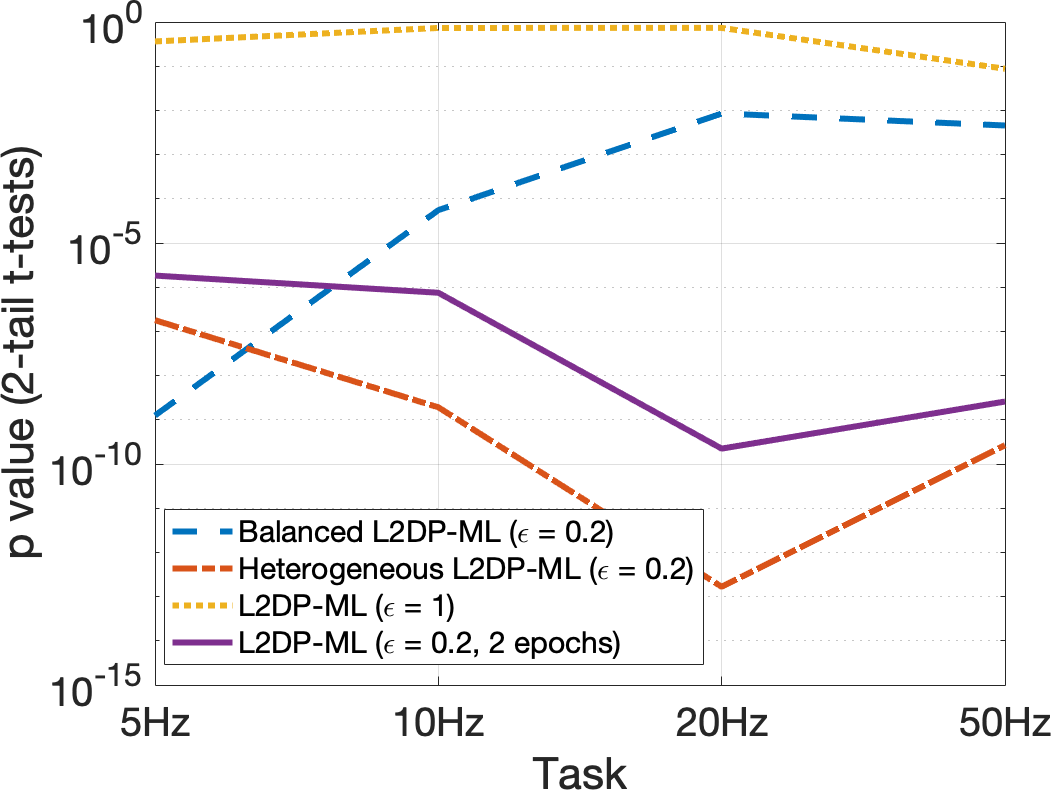}}
\caption{$p$ value for 2-tail t-tests on the (a) Permuted MNIST (20 tasks), b) Permuted CIFAR-10 (17 tasks), and (c) HARW (5Hz - 10Hz - 20Hz - 50Hz) (lower the better).}
\label{PvalueMNISTCIFAR}  
 \end{figure*}

 \begin{figure*}[h]
  \centering
\subfloat[]{\label{a}\includegraphics[scale=0.15]{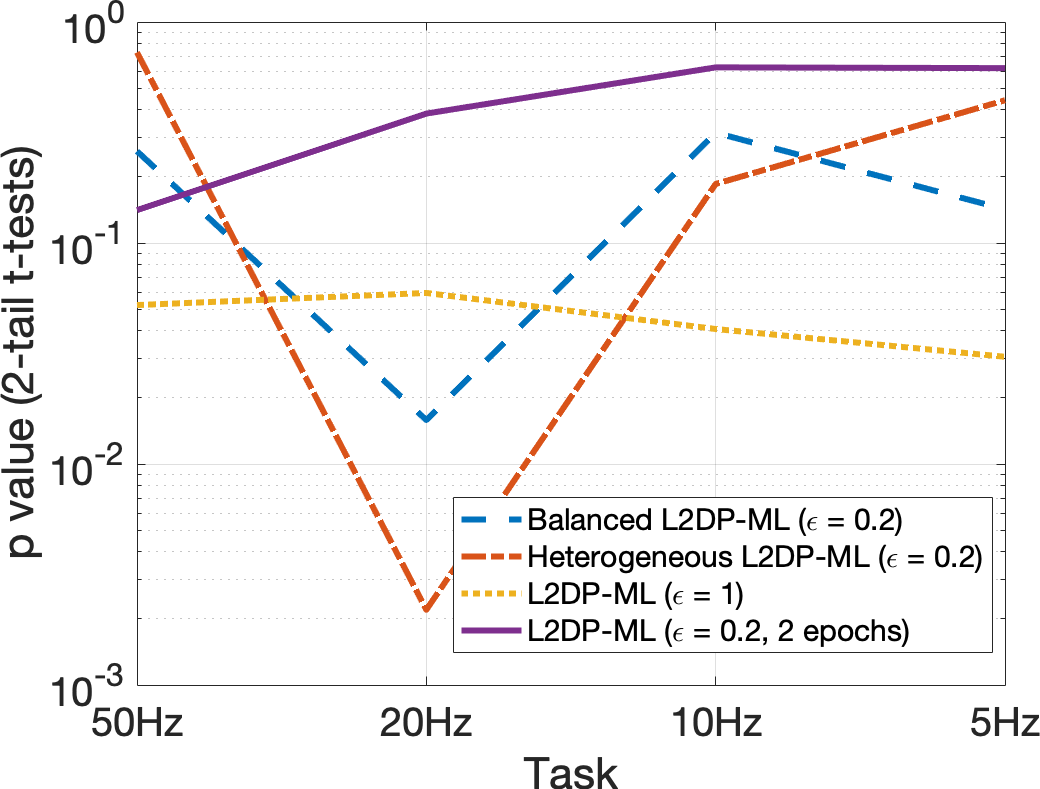}}\hspace*{0.1cm}
\subfloat[]{\label{b}\includegraphics[scale=0.15]{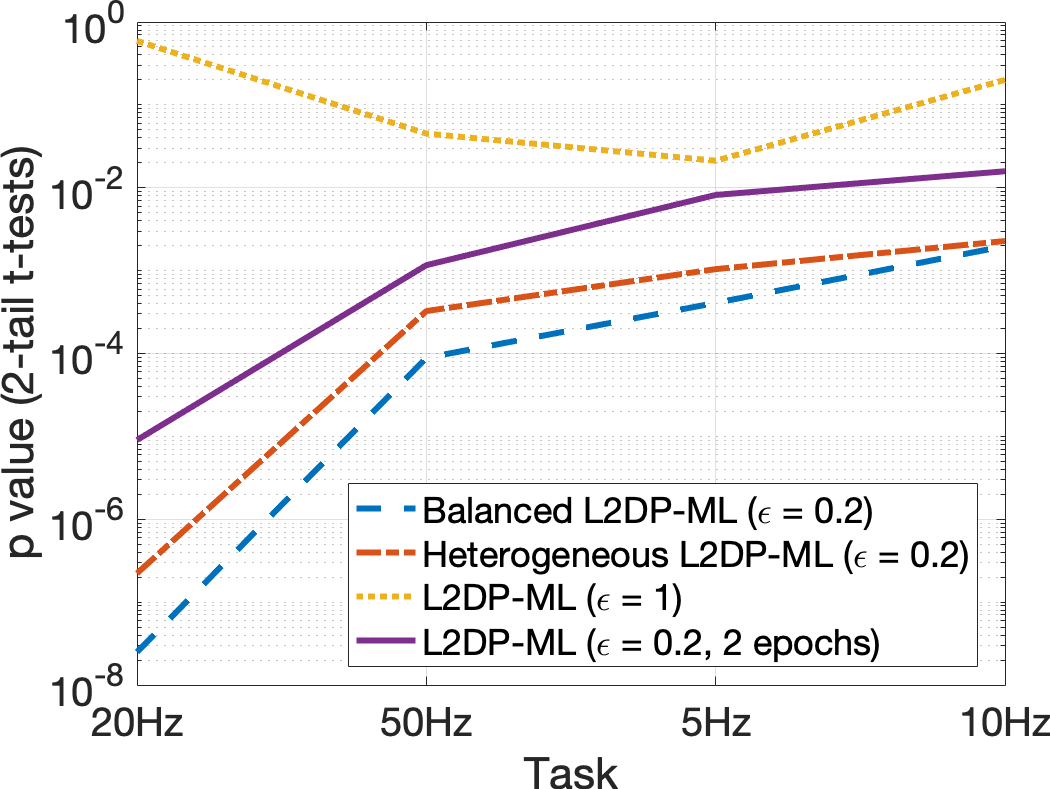}}
\hspace*{0.1cm}
\subfloat[]{\label{c}\includegraphics[scale=0.15]{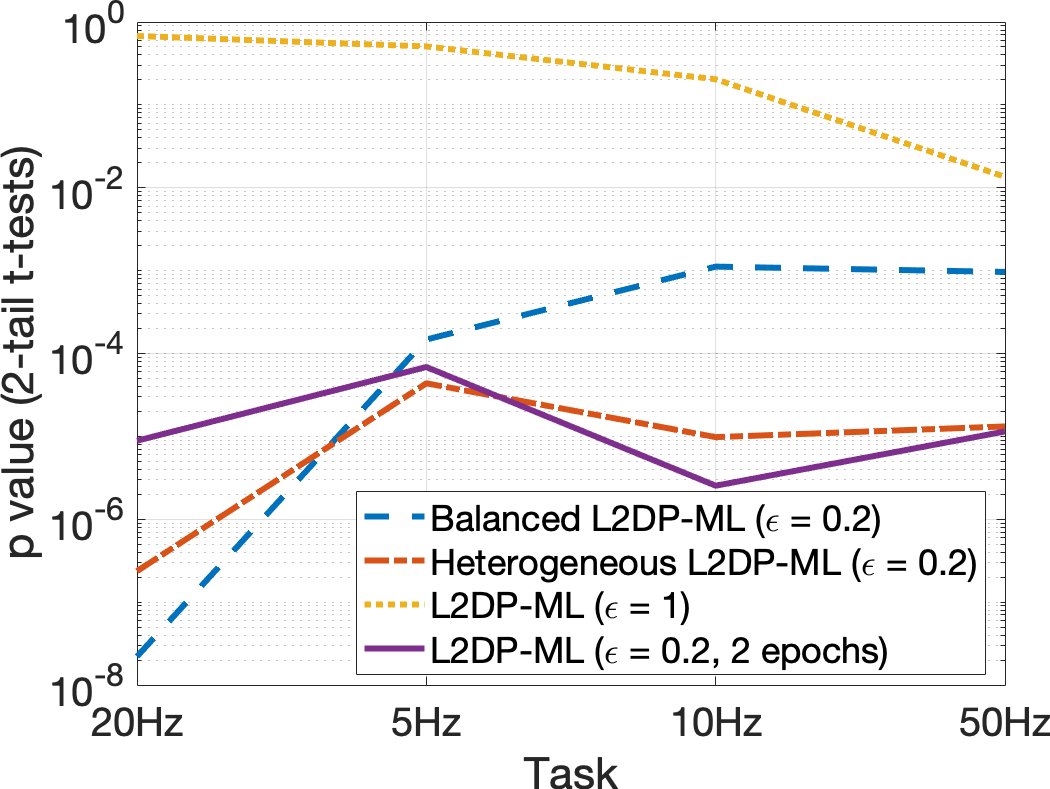}}
\caption{$p$ value for 2-tail t-tests on the HARW dataset with random task orders: (a) HARW 50Hz - 20Hz - 10Hz - 5Hz, (b) HARW 20Hz - 50Hz - 5Hz - 10Hz, and (c) HARW 20Hz - 5Hz - 10Hz - 50Hz. (lower the better).}
\label{HARExtraPValue}   
 \end{figure*}

 \begin{figure*}[t]
  \centering
\subfloat[]{\label{FigMNISTCIFARa2}\includegraphics[scale=0.2]{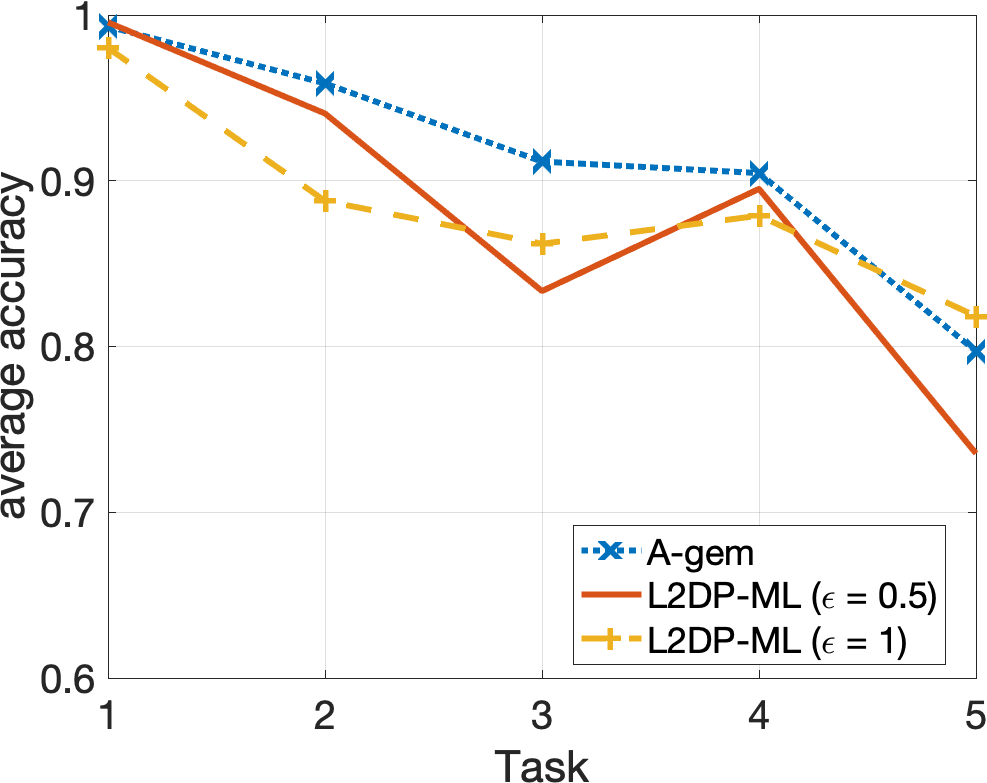}}\hspace*{2cm}
\subfloat[]{\label{FigMNISTCIFARb2}\includegraphics[scale=0.2]{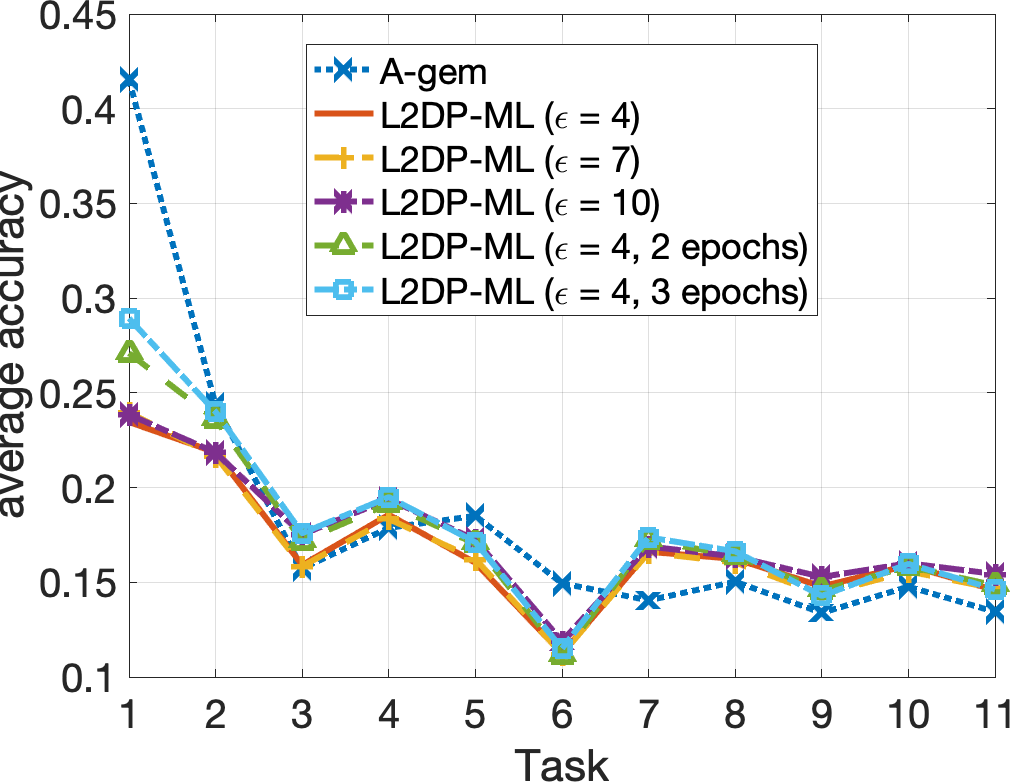}}
\caption{Average accuracy in the (a) Split MNIST (5 tasks), and b) Split CIFAR-10 and CIFAR-100 (11 tasks) (higher the better).}  
\label{FigMNISTCIFARCoLLas} 
 \end{figure*}

 \begin{table*}[!ht]
\caption{Average accuracy ($\%$) in hyper-parameter grid-search of NaiveGaussian mechanism given the permuted CIFAR-10 dataset.} 
\label{tab:cifar-10}
\begin{center}
\begin{tabular}{| c | c| c | c| c | }
\hline
 \multicolumn{2}{|c|}{\diagbox[height=5\line]{Privacy budget ($\epsilon$) \\ Noise scale ($z$) }{\\ \\ Clipping bound}}  & $0.01$ & $0.1$ & $1.0$ \\
\hline
\hline
 & $z = 2.5$  & $13.68$ &  $11.23$ &  $10.26$   \\
\cline{2-5} 
\multirow{5}*{\makecell{ $\epsilon =4.0$}} & $z = 2.4$  & $12.66$ &  $11.99$ &  $9.98$ \\
\cline{2-5} 
 & $z = 2.3$  & $11.56$ &  $11.40$ &  $10.09$ \\
\cline{2-5} 
& $z = 2.2$  & $13.79$ &  $11.99$ &  $10.30$ \\
\cline{2-5} 
& $z = 2.1$  & $13.50$ &  $11.39$ &  $10.11$ \\
\hline
& $z = 2.0$  & $15.12$ &  $12.94$ &  $10.26$   \\
\cline{2-5} 
\multirow{5}*{\makecell{ $\epsilon =7.0$}} & $z = 1.9$  & $14.67$ &  $12.39$ &  $10.34$ \\
\cline{2-5} 
& $z = 1.8$  & $14.32$ &  $11.79$ &  $10.28$ \\
\cline{2-5} 
& $z = 1.7$  & $15.26$ &  $12.55$ &  $11.33$ \\
\cline{2-5} 
& $z = 1.6$  & $14.64$ &  $12.28$ &  $11.04$ \\
\hline
  & $z = 1.5$  & $14.79$ &  $12.23$ &  $10.80$   \\
\cline{2-5} 
\multirow{5}*{\makecell{ $\epsilon =10.0$}} & $z = 1.4$  & $15.71$ &  $13.34$ &  $10.66$ \\
\cline{2-5} 
 & $z = 1.3$  & $15.12$ &  $12.96$ &  $11.49$ \\
\cline{2-5} 
& $z = 1.2$  & $14.65$ &  $12.05$ &  $10.64$ \\
\cline{2-5} 
& $z = 1.1$  & $11.42$ &  $11.15$ &  $10.14$ \\
\hline
\end{tabular}
\end{center}
\end{table*} 

\begin{table*}[t]
\centering
\caption{Average forgetting measure (smaller the better).}
\label{forgetting-appx}
\begin{tabular}{| c | c| c | c| c | c| c | c| c | c|}
\hline
\multicolumn{2}{|l|}{} & \textsc{L2DP-ML}  & A-gem \\
\hline
\multirow{2}*{\makecell{Split MNIST}}  &  $\epsilon = 0.5$ &   0.056 $\pm$	0.00324 & \multirow{2}*{0.195 $\pm$	0.00941}\\
\cline{2-3}
&  $\epsilon = 1$ & 0.019 $\pm$	0.00526  &   \\
\hline  
\multirow{5}*{\makecell{Split CIFAR-10/100}}  &  $\epsilon = 4$ &  0.027 $\pm$	0.00264  & \multirow{5}*{0.195 $\pm$	0.00688}\\
\cline{2-3}
& $\epsilon = 4$ (2 epochs) &  0.033 $\pm$	0.00276 &  \\
\cline{2-3}
&  $\epsilon = 4$ (3 epochs) &  0.046 $\pm$	0.00307 &  \\
\cline{2-3}
&  $\epsilon = 7$ &  0.027 $\pm$	0.00165 &  \\
\cline{2-3}
&  $\epsilon = 10$ &   0.021 $\pm$	0.00429 & \\
\hline 
\end{tabular}
\end{table*} 

\begin{table*}[t]
\caption{Average forgetting of the order of [20Hz, 5Hz, 10Hz, 50Hz] in the HARW task. (Smaller the better)}
\label{SuppHARTask}
\begin{center}
\resizebox{\textwidth}{!}{%
\begin{tabular}{| c | c|c|c|}
\hline 
 & L2DP-ML ($\epsilon = 0.2$) & L2DP-ML ($\epsilon = 0.5$) & L2DP-ML ($\epsilon = 1$) \\
 \cline{2-4}
 & 0.0928 $\pm$ 5.34e-5 & 0.0921 $\pm$ 8.64e-5 & 0.089 $\pm$ 8.64e-5 \\ 
 \cline{2-4}
HARW (20Hz - \text{ } & A-gem & Balanced A-gem & Balanced L2DP-ML ($\epsilon = 0.2$) \\
 \cline{2-4}
5Hz - 10Hz - 50Hz) & 0.0866 $\pm$ 1.1e-4 & 0.1723 $\pm$ 0.00066 & 0.144 $\pm$ 0.0031 \\  \cline{2-4}
 & L2DP-ML ($\epsilon = 0.2$, 2 epochs) & L2DP-ML ($\epsilon = 0.2$, 5 epochs) & Heterogeneous L2DP-ML ($\epsilon = 0.2$) \\ \cline{2-4}
 & 0.1161 $\pm$ 0.0003 & 0.1792 $\pm$ 0.0017 & 0.1395 $\pm$ 0.00026 \\
 \hline
\end{tabular}
}
\end{center} 
\end{table*}

\end{document}